\documentclass[journal]{IEEEtran}

\usepackage[utf8]{inputenc}
\usepackage[T1]{fontenc}
\usepackage{amsfonts,amsmath,amssymb}
\usepackage{anyfontsize}
\usepackage{booktabs}
\usepackage[inline]{enumitem}
\usepackage{graphicx}
\usepackage{mathtools}
\usepackage{microtype}
\usepackage{multirow}
\usepackage{subcaption}
\usepackage{url}
\usepackage[table]{xcolor}
\usepackage{xspace}
\usepackage[pagebackref]{hyperref}
\hypersetup{%
	pdfauthor={Taylor Mordan, Matthieu Cord, Patrick Pérez, Alexandre Alahi},%
	pdftitle={Detecting 32 Pedestrian Attributes for Autonomous Vehicles},%
	colorlinks=true,%
	linkcolor=blue,%
	citecolor=green%
}
\usepackage[nameinlink]{cleveref}

\newcommand{\eg}{e.g.\@\xspace}
\newcommand{\ie}{i.e.\@\xspace}

\begin{document}

\title{Detecting 32 Pedestrian Attributes\\for Autonomous Vehicles}

\author{Taylor~Mordan,
        Matthieu~Cord,
        Patrick~Pérez,
        and~Alexandre~Alahi
\thanks{T. Mordan and A. Alahi are with VITA, EPFL, Switzerland.}
\thanks{M. Cord is with Sorbonne Université, France.}
\thanks{M. Cord and P. Pérez are with valeo.ai, France.}}

\markboth{IEEE TRANSACTIONS ON INTELLIGENT TRANSPORTATION SYSTEMS}%
{Mordan \MakeLowercase{et al.}: Detecting 32 Pedestrian Attributes for Autonomous Vehicles}

\maketitle

\begin{abstract}
Pedestrians are arguably one of the most safety-critical road users to consider for autonomous vehicles in urban areas.
In this paper, we address the problem of jointly detecting pedestrians and recognizing 32 pedestrian attributes from a single image.
These encompass visual appearance and behavior, and also include the forecasting of road crossing, which is a main safety concern.
For this, we introduce a Multi-Task Learning (MTL) model relying on a composite field framework, which achieves both goals in an efficient way.
Each field spatially locates pedestrian instances and aggregates attribute predictions over them.
This formulation naturally leverages spatial context, making it well suited to low resolution scenarios such as autonomous driving.
By increasing the number of attributes jointly learned, we highlight an issue related to the scales of gradients, which arises in MTL with numerous tasks.
We solve it by normalizing the gradients coming from different objective functions when they join at the fork in the network architecture during the backward pass, referred to as fork-normalization.
Experimental validation is performed on JAAD, a dataset providing numerous attributes for pedestrian analysis from autonomous vehicles, and shows competitive detection and attribute recognition results, as well as a more stable MTL training.
\end{abstract}


\section{Introduction}

\IEEEPARstart{A}{lthough} autonomous vehicles have already demonstrated successful autonomy on highways~\cite{arnold2019survey, claussmann2019review, kuutti2019survey}, urban areas and cities remain a challenge due to a higher degree of diversity in situations and actors.
Pedestrians are arguably one of the most important categories to consider in this context, as they are more mobile and less predictable than vehicles.
As part of the Vulnerable Road Users (VRUs), they have recently received more attention from the autonomous vehicle community on multiple related tasks~\cite{ridel2018a, ferranti2019safevru, rasouli2019autonomous, alvarez2020autonomous}, which is essential to guarantee safety.

In the context of autonomous vehicles, lots of cues about pedestrians are available (\eg, intentions to cross the road, eye contacts with the driver, gestures for communication, descriptions of their appearances) on some datasets~\cite{rasouli2017are, rasouli2019pie}.
Human attributes have proven to be effective intermediate features to combine with other tasks to improve results~\cite{tian2015pedestrian, zhang2019attribute, sulistiyo2020attribute} or to learn more generic representations~\cite{chen2012describing, deng2014pedestrian, li2019a}.
It opens the way for Multi-Task Learning (MTL)~\cite{caruana1997multitask, zamir2018taskonomy} to exploit commonalities between these attributes, and to meet strict time and memory requirements of real-world autonomous vehicles by sharing computational resources between attributes.

However, it is difficult to learn a large number of tasks simultaneously, or to learn with heterogeneous annotations, where each example is only labeled for a subset of the tasks~\cite{kokkinos2017ubernet}.
Most previous works addressing attribute recognition in an autonomous vehicle context~\cite{volz2015feature, rasouli2017are, varytimidis2018action, rasouli2019pedestrian} have only learned a limited number of attributes, missing out on some benefits of MTL.
In addition, they often directly use ground-truth pedestrian bounding boxes as input, then relying on a prior pedestrian detector in real-world applications.
Such results, assuming a perfect detector, are therefore optimistic~\cite{saleh2019real}.

\begin{figure*}[t]
  \centering
  \includegraphics[width=\linewidth]{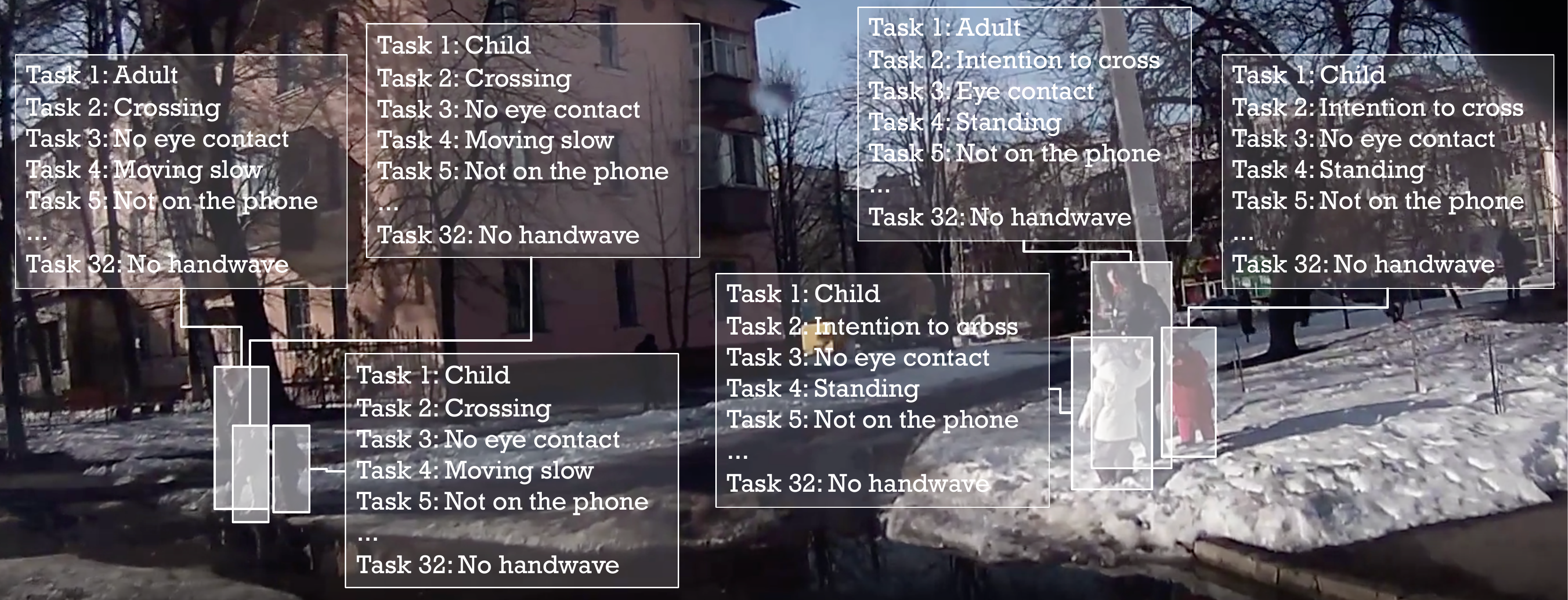}
  \caption{\textbf{Illustration of joint pedestrian detection and attribute recognition} based on ground truth annotations from dataset JAAD~\cite{rasouli2017are}. We present a Multi-Task Learning (MTL) approach that jointly detects pedestrians and recognizes 32 pedestrian attributes, encompassing appearance, behavioral cues, and forecasting cues such as intention to cross the road.}
  \label{fig:pull}
\end{figure*}

In this paper, we first propose an end-to-end MTL approach, performing both detection of pedestrians and attribute recognition for all of them in a unified way based on bottom-up composite fields~\cite{kreiss2019pifpaf}, as illustrated in \Cref{fig:pull}.
Given a single image, we jointly detect pedestrians and their visual appearances, behavioral attributes and even intentions.\footnote{While the focus of our work is to push the limits of MTL given a single image, future work can use temporal information (multiple frames) to recognize behavioral and intentional attributes.}
Field formalism has proven to be effective for urban scenes, especially at low resolution, which is often the case for pedestrians imaged from a vehicle.
Performing detection and attribute recognition jointly should be more computationally efficient than the two-stage pipelines implicitly implied by the previous works, and enables the evaluation of the whole perception step for a more representative understanding of the system.
Secondly, we greatly increase the number of attributes to have the most informative and complete description of pedestrians, and to analyze how the model behaves when scaling MTL to numerous tasks.
In this scenario, we highlight an issue with merging gradients from different tasks, and propose fork-normalization, a simple yet effective solution to handle it and improve training by normalizing gradients during back-propagation.
Finally, we experimentally validate our findings on the Joint Attention in Autonomous Driving (JAAD) dataset, a dataset designed around pedestrians in an autonomous vehicle context.
Leveraging all the attributes annotated on this dataset with enough examples for proper training, we work with up to 32 attributes.
A complete breakdown of the attributes used is given in \Cref{sec:exp_setup}.\footnote{Code available at \url{https://github.com/vita-epfl/detection-attributes-fields}.}

\section{Related Work}

\paragraph{Pedestrian Analysis from Vehicles}
Understanding pedestrians' behaviors around vehicles is a major milestone for safety in urban areas~\cite{ridel2018a, rasouli2019autonomous, malla2020titan}.
One critical interaction is road crossing~\cite{volz2016a, rasouli2019pedestrian, saleh2019real}, for which several large-scale datasets have recently been released to enable data-driven approaches~\cite{rasouli2017are, rasouli2019pie, liu2020spatiotemporal}.
These datasets annotate pedestrians interacting with the ego-vehicle with information on whether they cross the road in front of it, along with a varying amount of visual features.
Trajectory prediction is another main goal of pedestrian analysis~\cite{coscia2016point, alahi2017learning, saleh2017intent, rasouli2019pie, ridel2019understanding, kothari2020human}.
The most successful approaches usually learn temporal dynamics with LSTM networks and focus on interactions between agents to predict plausible trajectories.
Alternative approaches jointly solve both problems by finding the short-term destinations and check whether it is on the other side of the road~\cite{rehder2015goal, rehder2018pedestrian}, or by estimating future frames and predicting target tasks from them~\cite{chaabane2019looking, gujjar2019classifying, bouhsain2020pedestrian}.

\paragraph{Human Attributes}
Describing humans with a set of attributes~\cite{gkioxari2015actions, gkioxari2015contextual, li2016human, tang2019improving} has recently received attention.
The common approach for these methods is mainly to focus on finding different human parts or regions of bounding boxes that are directly linked to attributes, and compute features associated with them to recognize attributes.
This has been shown to lead to features that transfer well to other tasks~\cite{chen2012describing, li2019a} and datasets~\cite{deng2014pedestrian}.
In particular, attribute-based representations are often used for person re-identification~\cite{layne2012person, schumann2017person, lin2019improving}.
In this application, visual appearances can change significantly between multiple views of the same person, due to different viewpoints or lighting conditions.
These works therefore first extract high-level attributes that should remain constant for all instances of the same people, for example about clothing, and then use them to match identities of examples.
Regarding pedestrians, it has been shown that attributes have a direct impact on detection performances~\cite{rasouli2018it}.
In the context of autonomous vehicles, they have been learned~\cite{rasouli2017are, varytimidis2018action, pop2019multi} and used to support other tasks, \eg, road crossing prediction~\cite{kohler2012early, bonnin2014pedestrian, volz2015feature, schneemann2016context, rasouli2017are, varytimidis2018action}, detection~\cite{tian2015pedestrian, zhang2019attribute}, and segmentation~\cite{sulistiyo2020attribute}.
In particular, pose is regularly studied~\cite{quintero2015pedestrian, fang2017on, fang2018is, wang2019estimating}.
Approaches usually either leverage body languages, through poses or by focusing on head and leg regions, to forecast future behaviors of pedestrians, or describe appearances of both pedestrians and surrounding scenes to better understand the contexts.

Similarly to our work, Tian et al.~\cite{tian2015pedestrian} have also explored doing joint pedestrian detection and attribute recognition.
However, detection is done using a Sliding Window approach, which is more time-consuming than our single network forward pass.
They also learn some attributes in a multi-task fashion, but we further develop this approach by learning around three times more attributes and introducing a normalization in back-propagation to solve gradient scale issues during training.

\paragraph{Multi-Task Learning}
The goal of MTL~\cite{caruana1997multitask} is to share learning capacity among several tasks to benefit from transfer between them~\cite{zamir2018taskonomy} and to reduce computational requirements.
It has been successfully applied to a driving context~\cite{chabot2017deep}.
However, experimentation with larger numbers of tasks has shown that boosts in performances are not systematic with enough tasks~\cite{kokkinos2017ubernet}, and that the system has to be optimized for the specific target objectives to yield optimal transfer~\cite{bell2020groknet}.
It has also be shown that MTL helps providing robustness against adversarial attacks~\cite{mao2020multitask}, which is an important safety concern for autonomous vehicles.
Most works can be divided into two categories, depending on the level at which they affect interactions among objectives.
Some approaches optimize performances by learning the network structures~\cite{misra2016cross, lu2017fully} or the amount of sharing between tasks~\cite{yang2017deep, meyerson2018beyond, liu2019end, vandenhende2020mti}.
Others work with any given architecture and learn loss weights to balance tasks~\cite{chen2018gradnorm, guo2018dynamic, kendall2018multi} or manipulate gradients to reduce negative transfer~\cite{suteu2019regularizing, lu202012, yu2020gradient}.
These latter ones particularly relate to our approach, as they look into gradient manipulations to ensure stability during learning, studying vector angles or norms.
Intuitively, the norm of the sum of gradients should be influenced by the angles between gradients.

This paper focuses on learning a large number of tasks in a driving context, where labels are often available for a subset of tasks only, and deals with gradient scale issues for a generic MTL architecture.

\begin{figure*}[t]
  \centering
  \includegraphics[width=\linewidth]{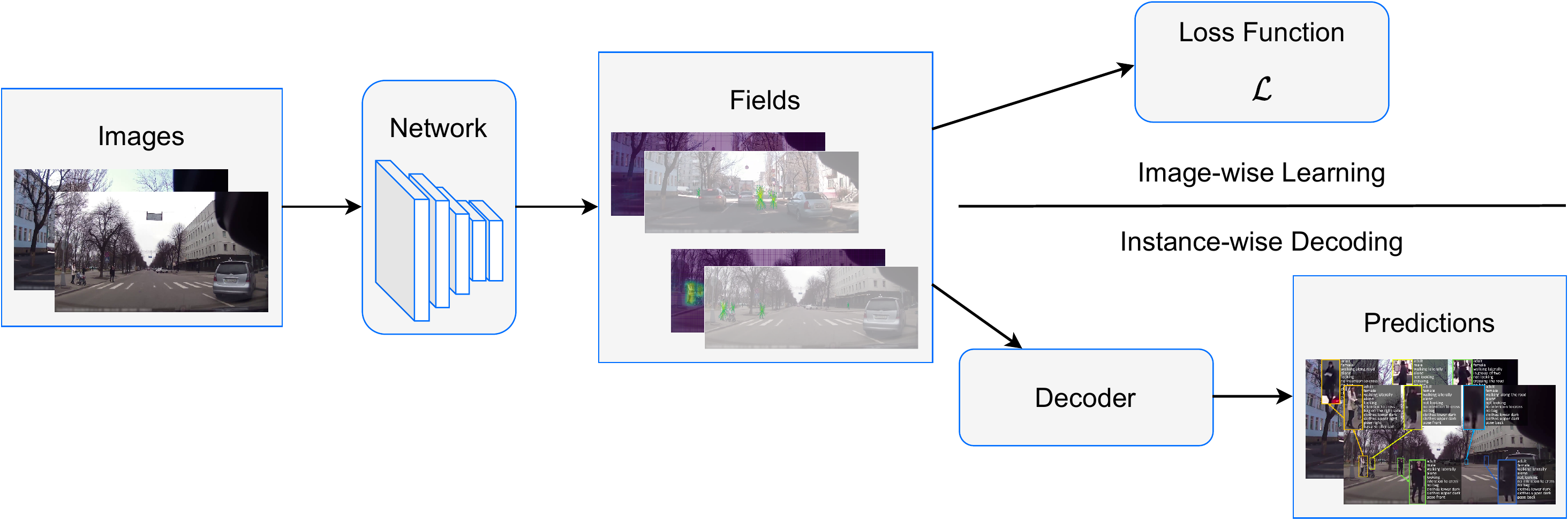}
  \caption{\textbf{Data flow diagram.} The multi-task network produces fields from images. During training, each field is learned with an associated image-wise loss function. At test-time, all the fields are decoded together to yield a set of instance predictions.}
  \label{fig:dataflow}
\end{figure*}

\section{Joint Pedestrian Detection and Attribute Recognition}
\label{sec:model}

\begin{figure*}
    \centering
    \begin{subfigure}[b]{\textwidth}
        \centering
        \includegraphics[width=\textwidth]{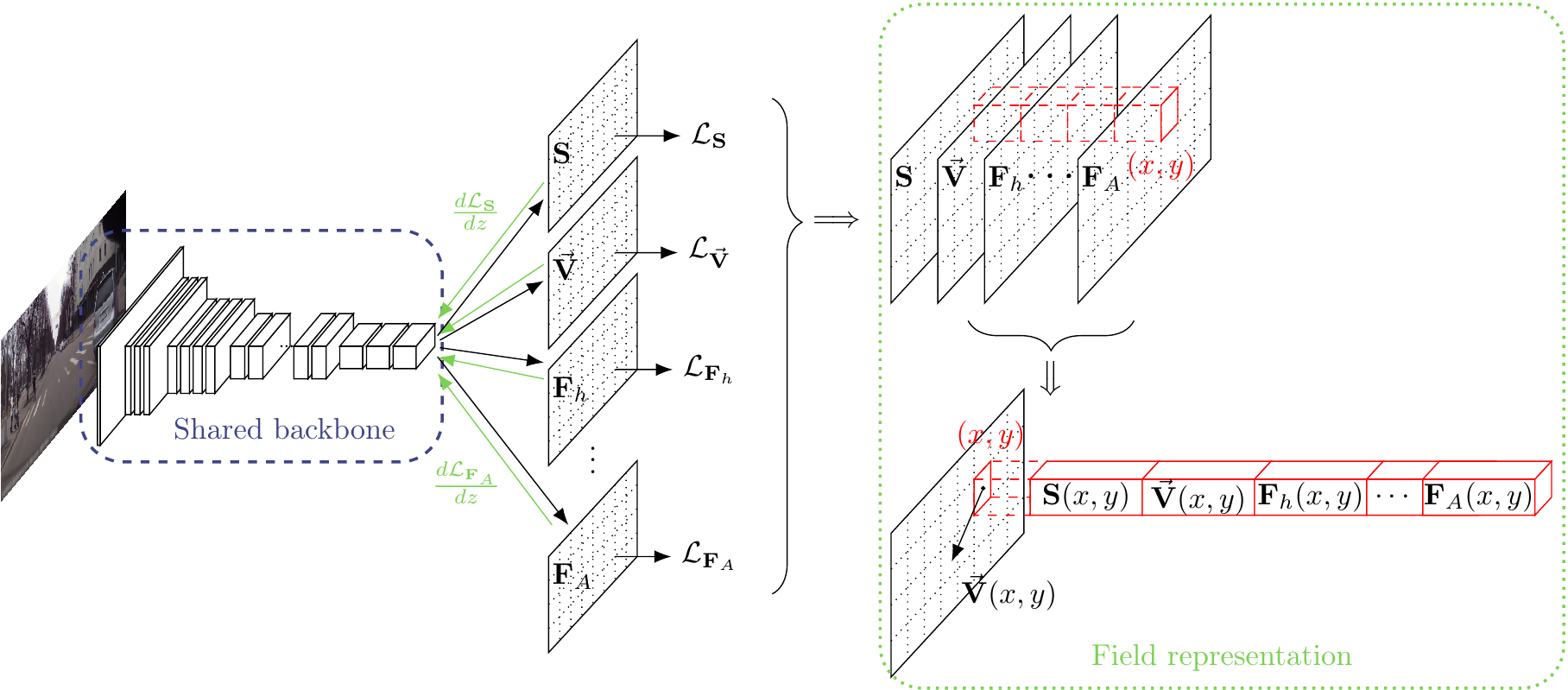}
        \caption{Architecture of our model}
    \end{subfigure}
    \par\medskip
    \null\hfill
    \begin{subfigure}[b]{.48\textwidth}
        \centering
        \includegraphics[width=\textwidth]{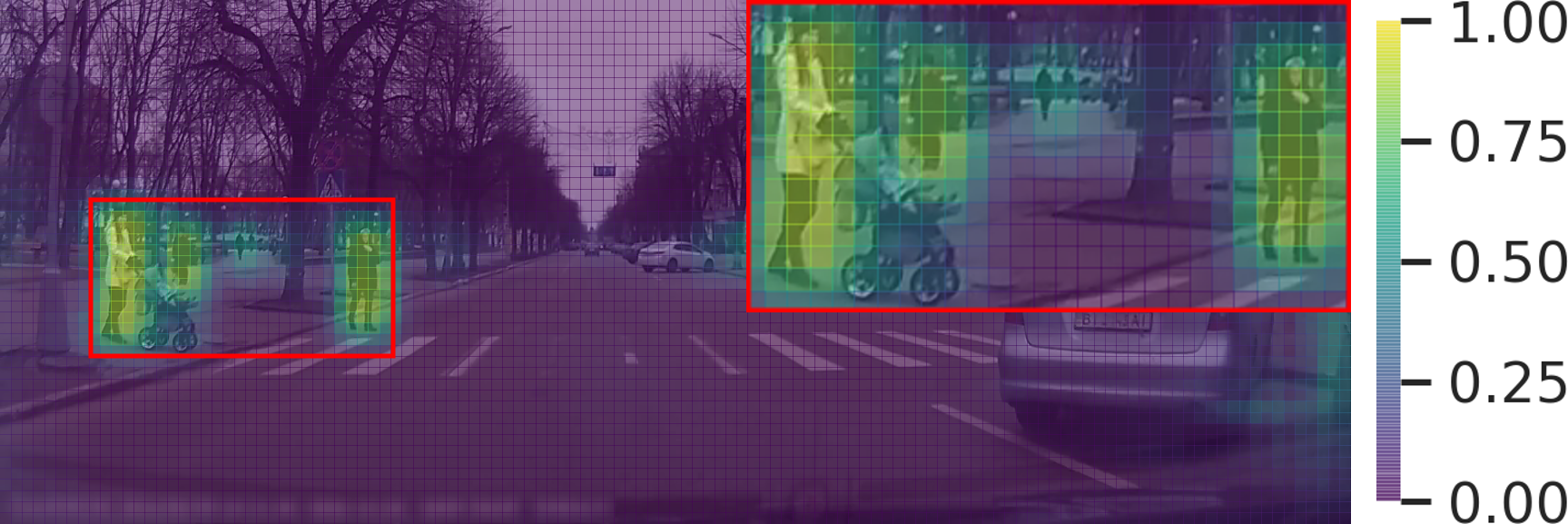}
        \caption{Field $\mathbf{S}$}
    \end{subfigure}
    \hfill\hfill
    \begin{subfigure}[b]{.48\textwidth}
        \centering
        \includegraphics[width=\textwidth]{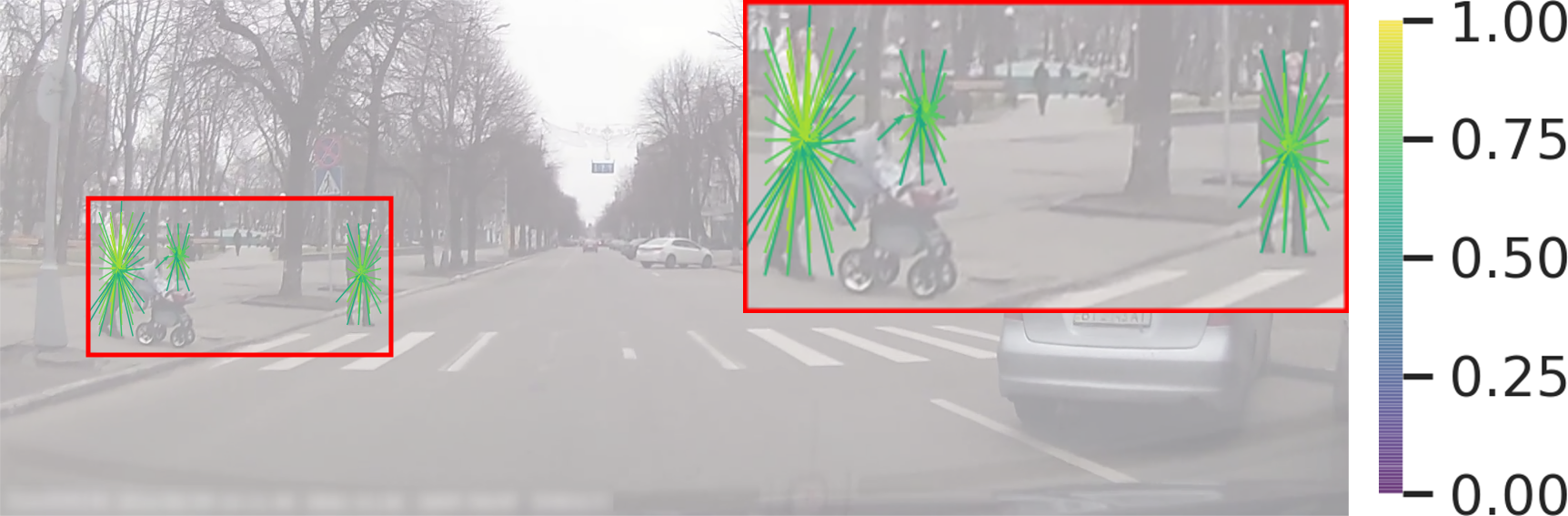}
        \caption{Field $\vec{\mathbf{V}}$}
    \end{subfigure}
    \hfill\null
    \caption{\textbf{Field-based representations.} (a) The model is composed of a backbone branching into a task-specific field for each task. Detection of pedestrian instances is done with fields $\mathbf{S}$ and $\vec{\mathbf{V}}$, with $\mathbf{F}_h$ and $\mathbf{F}_w$ for the bounding boxes' dimensions. The recognition of other attributes is done with additional fields $\mathbf{F}_1, \ldots, \mathbf{F}_A$ for $A$ attributes. When considering all fields together, at each cell $(x,y)$ in the model's output, there is a local representation localizing the center of the pedestrian it belongs to, along with predictions for all attributes. During inference (not shown in the figure), all the local attributes associated with a same pedestrian are aggregated through a vote to yield a unique prediction per detection. (b-c) Example fields $\mathbf{S}$ and $\vec{\mathbf{V}}$, with top right windows zooming in on pedestrians for more details. The color code for the heatmap from $\mathbf{S}$ and the arrows from $\vec{\mathbf{V}}$ is based on the confidence scores $\operatorname{Sigmoid}\left(\mathbf{S}(x,y)\right)$ of the cells.}
    \label{fig:model}
\end{figure*}

Our model is based on the common Multi-Task Learning (MTL) network architecture consisting of a single shared backbone followed by a separate predictor for each task.
As shown in \Cref{fig:dataflow}, we rely on the composite field formalism~\cite{kreiss2019pifpaf}, where the model's output is a set of spatial feature maps encoding both the locations and attributes of all the pedestrians in the scene.
Learning is done on image-wise fields, and predictions are obtained with a post-processing instance-wise decoding step.
Detection and attribute recognition tasks are here combined in a unified approach, naturally lending itself to MTL~\cite{tian2015pedestrian, zhang2019attribute}.

\subsection{Field Formalism}

For every pedestrian attribute $f$ we want to predict (\eg, location, bounding box dimension, visual or behavioral attribute), we consider an associated field $\mathbf{F}_f$ (for clarity in the following, we will omit the subscript $f$ when it is clear from the context), which is a spatial map on whole images and whose cells $\mathbf{F}(x,y)$ locally encode an estimation of the given attribute $f$ based on the neighborhoods around the locations $(x,y)$.

We consider two kinds of fields.
Scalar fields $\mathbf{F}$ encode global attributes $f$ about pedestrians from all their respective cells regardless of their relative locations (see \Cref{fig:model}~\hyperlink{fig:model}{(b)} for an example).
These attributes can be binary (\eg, intention to cross the road), categorical (\eg, age category), or continuous (\eg, height).
In case of a categorical attribute, the field $\mathbf{F}$ has as many channels as classes for the attribute, and $\mathbf{F}(x,y)$ contains predictions for all these classes.
A vectorial field $\vec{\mathbf{F}}$ points to spatial locations $\vec{f}=(x^{f},y^{f})$ relative to the pedestrians' positions, with vectors pointing from all their respective cells to the desired locations (see \Cref{fig:model}~\hyperlink{fig:model}{(c)} for an example).
The only vectorial attribute we have in this paper is the locations of the instances' centers, but this formulation could be used with any localized attribute, such as pose keypoints for example.

As depicted in \Cref{fig:model}~\hyperlink{fig:model}{(a)}, when taking all such fields from the model stacked together, every cell $(x,y)$ of the output contains a complete representation around this location.
This includes an estimation of the presence of a pedestrian, and in the case of a detected pedestrian, the associated location and local predictions for all attributes.
When no pedestrian is detected at the location, the attribute predictions will not be used and do not affect the results.

This image-wise view based on fields is later converted into a set of instance-wise predictions with a post-processing decoding step, as shown in \Cref{fig:dataflow}: all cells $(x,y)$ associated with the same pedestrians are aggregated to yield a single prediction $f_p$ or $\vec{f}_p$ from field $\mathbf{F}$ or $\vec{\mathbf{F}}$ for detected pedestrian $p$ (see Section \ref{sec:decoding}).
These spatial formulation and inference implicitly take context into consideration and are therefore well suited to low resolution contexts~\cite{kreiss2019pifpaf}.

\subsection{Detection and Attribute Recognition}

Pedestrian detection is achieved with multiple fields.
First, field $\mathbf{S}$ (see \Cref{fig:model}~\hyperlink{fig:model}{(b)}) estimates how likely cells $(x,y)$ are to belong to pedestrian instances.
This is done through confidence scores $\operatorname{Sigmoid}\left(\mathbf{S}(x,y)\right) \in \left[0,1\right]$: a value close to 1 indicates a pedestrian, while a value close to 0 indicates background.
Then, field $\vec{\mathbf{V}}$ (see \Cref{fig:model}~\hyperlink{fig:model}{(c)}) is used to localize the centers of the pedestrian instances, \ie, the centers of their bounding boxes.
At any cell $(x,y)$ belonging to a pedestrian, the vector $\vec{\mathbf{V}}(x,y)$ should point from the cell to the corresponding pedestrian's center.
The two fields $\mathbf{S}$ and $\vec{\mathbf{V}}$ are used together to detect all pedestrians present in the images and spatially cluster cells into separate instances in a bottom-up way~\cite{kendall2018multi}.
This does not require any hand-crafted modules that are common in detection pipelines, such as prior boxes or non-maximum suppression step, which is itself a main motivation of recent object detection methods \cite{zhou2019objects, carion2020end}.

However, the fields $\mathbf{S}$ and $\vec{\mathbf{V}}$ only allow for point detection, \ie, detecting pedestrians by their central point.
Regular box detection is obtained by adding two other scalar fields, for the heights and widths of the bounding boxes.
Grouped together, these form the first attribute we consider for pedestrians: the bounding box attribute.
Note that although the field $\mathbf{S}$ already provides a coarse segmentation of the pedestrians in the feature space, this is not accurate in the image space due to the large stride of the network.
Also, getting the bounding box as a field prediction should be more robust, especially in the case of occlusion, as it is averaged over multiple cells rather than relying on the detection of cells at the edges of the pedestrians.

Once all cells are attributed to specific pedestrian instances or background, attribute recognition is straightforward.
Given an attribute $f$ (including height and width) or $\vec{f}$ (including the center of the detection) with associated field $\mathbf{F}$ or $\vec{\mathbf{F}}$, a prediction for a pedestrian instance is obtained by a vote of all values $\mathbf{F}(x,y)$ or $\vec{\mathbf{F}}(x,y)$ whose cells $(x,y)$ have been clustered in this instance.

\subsection{Image-wise Learning}
\label{sec:field_learning}

We consider annotations for the tasks of detection and attribute recognition under the form of a bounding box for each pedestrian, along with a list of attributes.
The attributes can be scalar (binary, categorical or continuous values) or vectorial (spatial coordinates).
Note that pedestrians are not all annotated for all the attributes, depending on what is visible in the data, making the set of attributes available for each pedestrian different.
Missing attributes are not used during training (no loss back-propagated for the associated tasks), and predictions for them are not evaluated at inference.

Learning is carried out with image-wise loss functions, independently of the number of pedestrians present in the scene.
Binary and categorical fields can be learned with any classification loss function, and continuous and vectorial fields with any regression loss function.
Binary targets for the detection field $\operatorname{Sigmoid}\left(\mathbf{S}\right)$ are set to 1 for all cells $(x,y)$ belonging to pedestrians, defined as $(x,y)$ being within a pedestrian's bounding box, and to 0 if they correspond to background.
For all other fields, there are no targets defined for background cells and no loss is back-propagated there.
Regarding a pedestrian cell $(x,y)$, all fields learn the attributes of the corresponding pedestrian.
Scalar fields $\mathbf{F}$ first go through activation functions $\operatorname{act}$ to have the right output ranges: $\operatorname{Sigmoid}$ for binary attributes, $\operatorname{Softmax}$ for categorical attributes and $\operatorname{Identity}$ for continuous attributes.
Targets are then set to the values of the corresponding pedestrian's attributes $f$, duplicated over all the cells within the bounding box.
For vectorial fields $\vec{\mathbf{F}}$, the targets are the vectors pointing from that cell $(x,y)$ to the locations $\vec{f}=(x^{f},y^{f})$ of the attributes, \ie, $\vec{f} - (x,y)$.
Note that the only vectorial field used in this paper is $\vec{\mathbf{V}}$, but the setup would be the same for any other such field.

\begin{figure*}[t]
	\begin{center}
		\includegraphics[height=3.85cm]{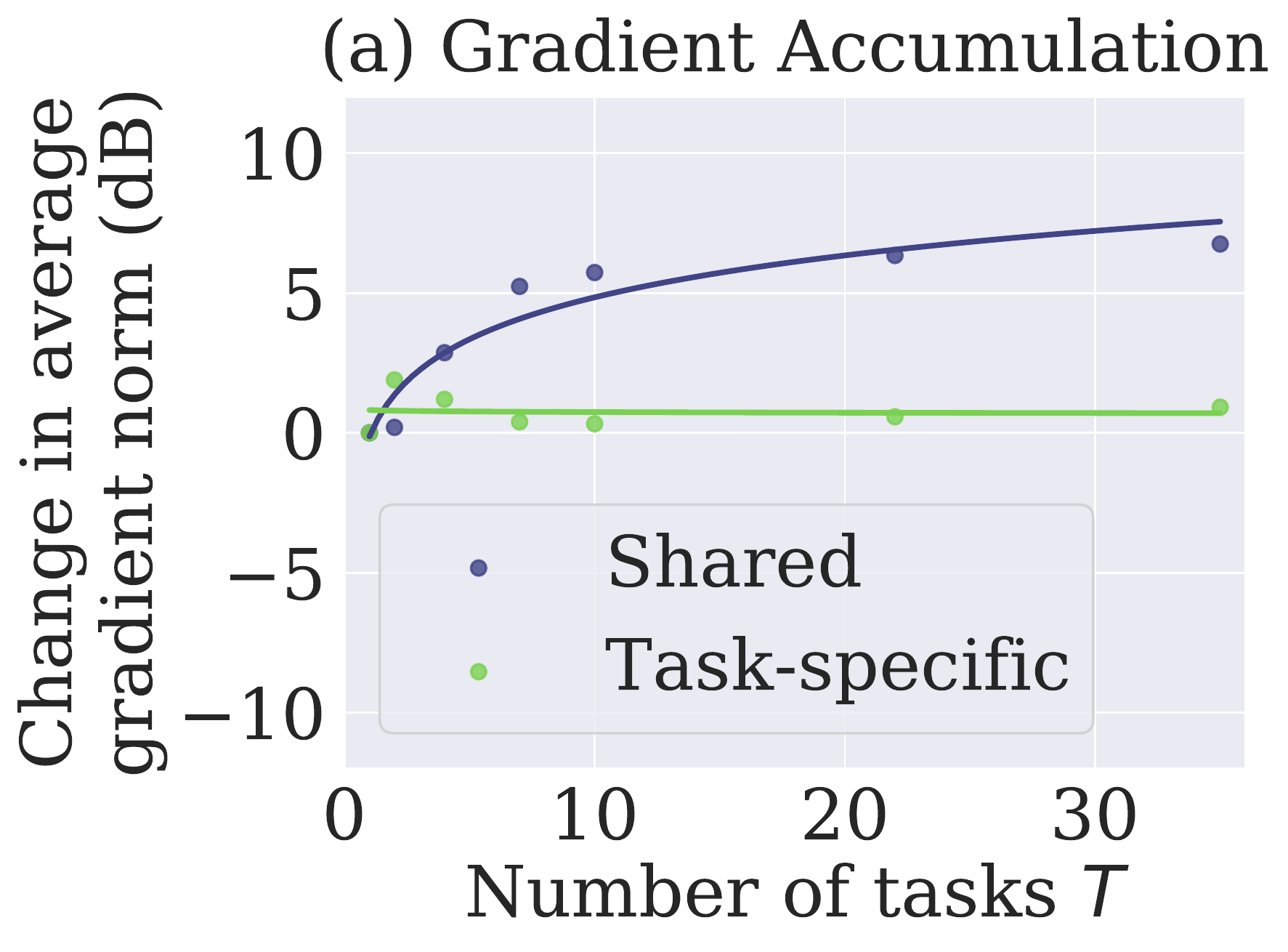}
		\hfill
		\includegraphics[height=3.85cm]{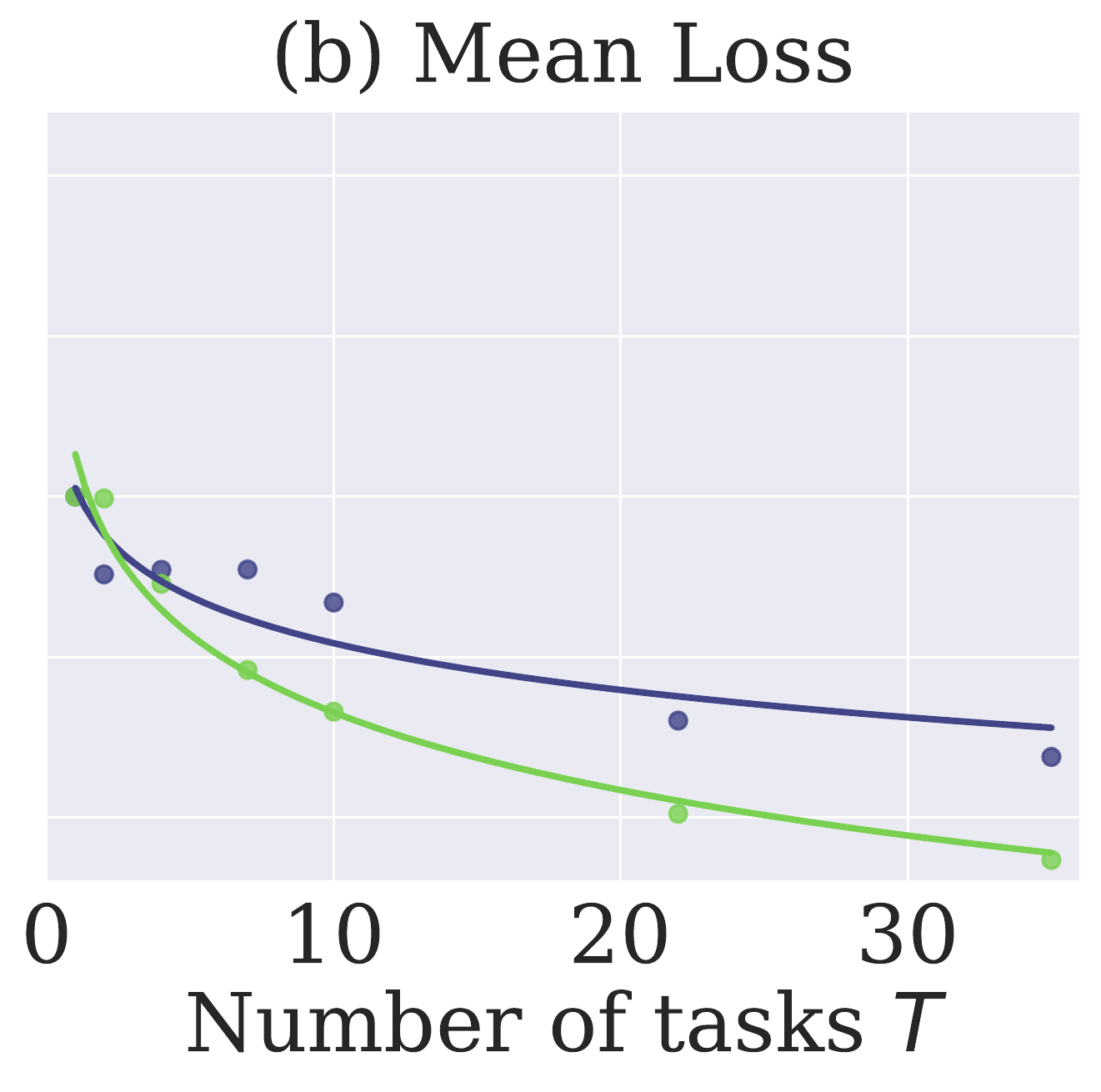}
		\hfill
		\includegraphics[height=3.85cm]{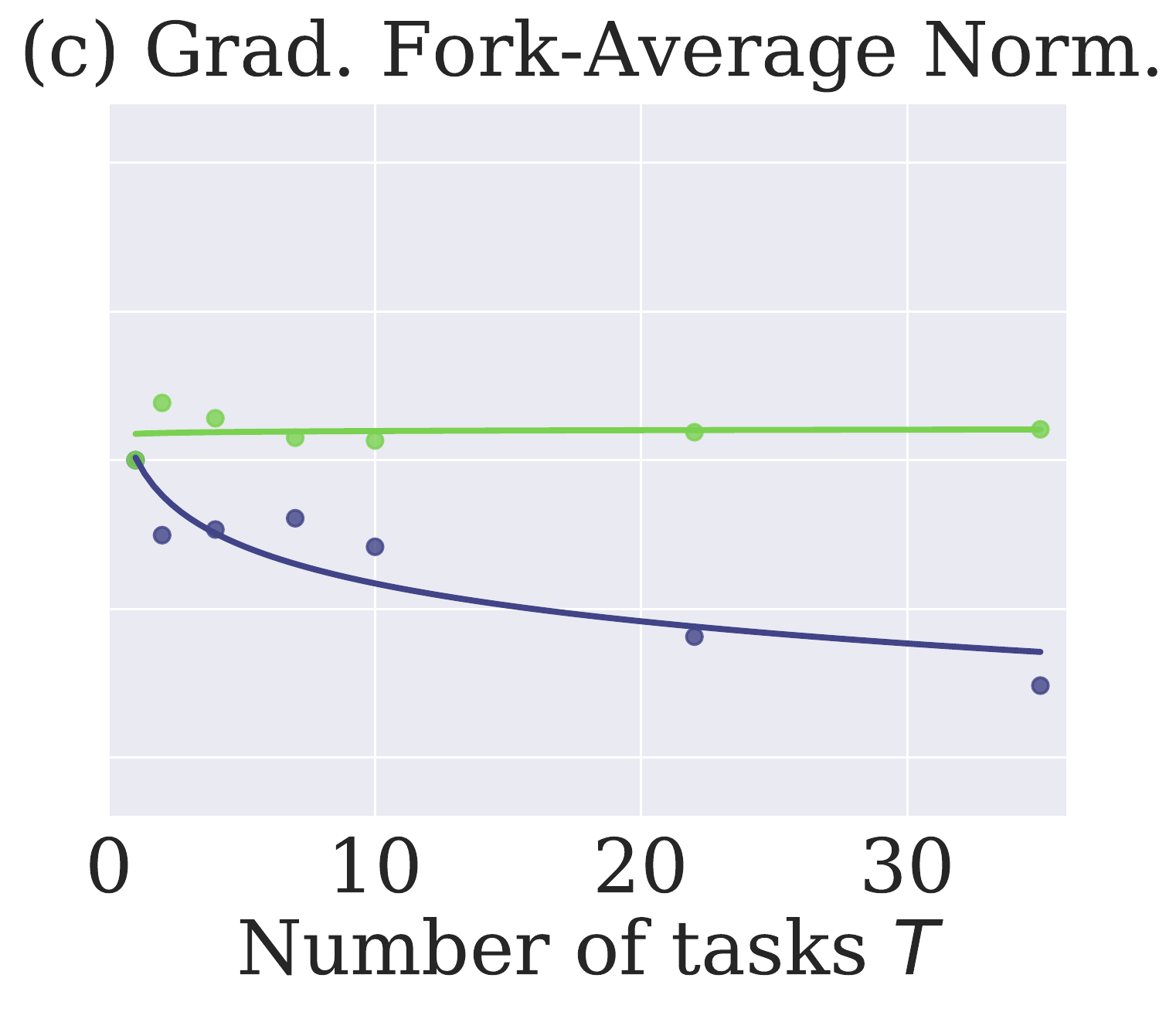}
		\hfill
		\includegraphics[height=3.85cm]{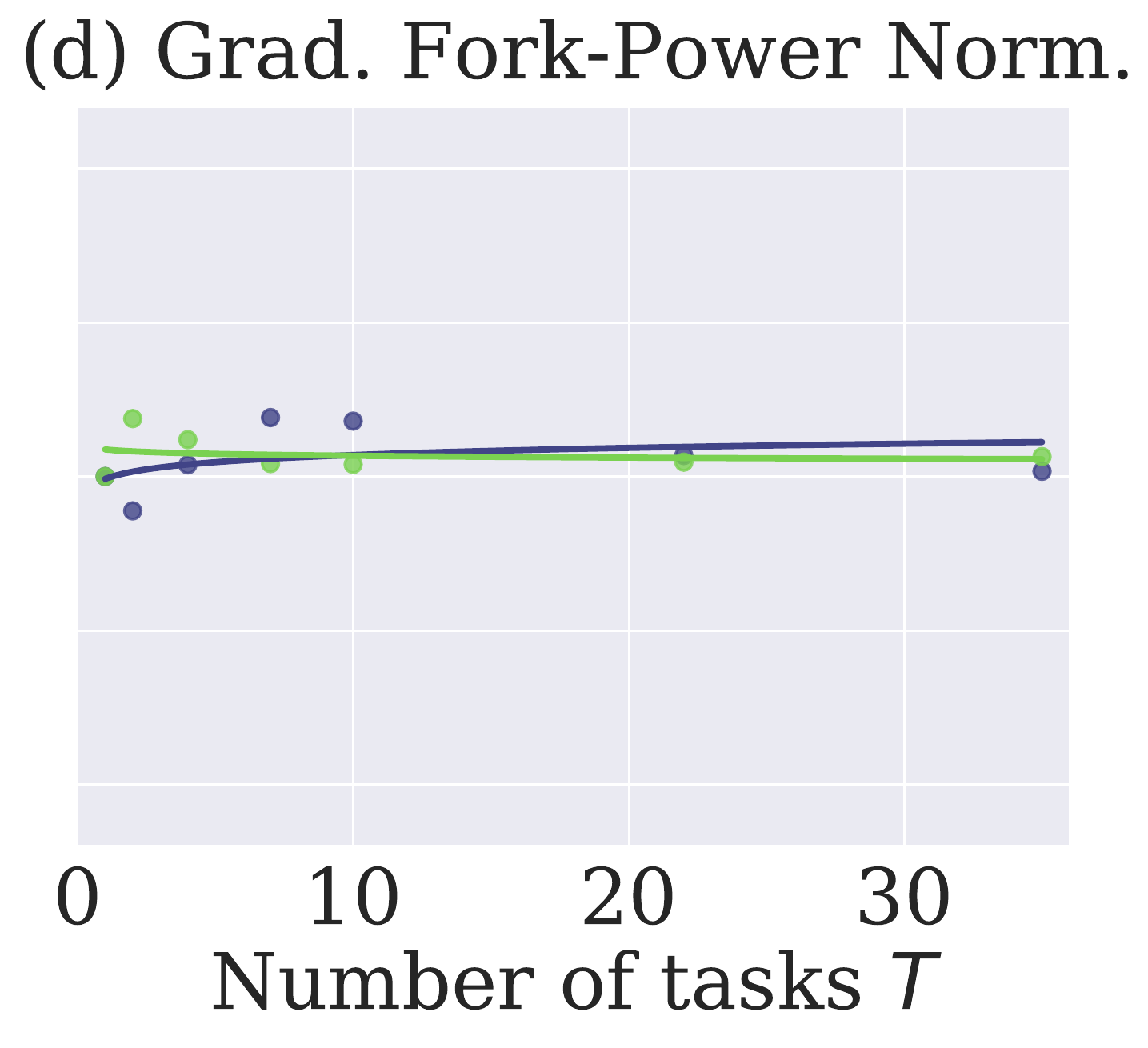}
	\end{center}
	\hypertarget{fig:grad_norm}{}%
	\caption{\textbf{Relative change in average gradient norm} (in dB), in both the shared backbone and task-specific sub-networks, when increasing the number of tasks $T$ learned with (a) standard gradient accumulation; (b) global mean of loss functions; (c) our proposed gradient fork-average normalization; (d) our proposed gradient fork-power normalization. Norm values are the average norms of gradients over the first epochs. Points are actual measurements and lines are linear fits in log-log spaces.}
	\label{fig:grad_norm}
\end{figure*}

\subsection{Instance-wise Decoding}
\label{sec:decoding}

As displayed in \Cref{fig:dataflow}, the model's output is a set of fields on whole images.
Predictions are obtained from this image representation in a bottom-up way, with a post-processing step not requiring training.
Pedestrian detection is first done by clustering cells into a set of instances, using fields $\mathbf{S}$ and $\vec{\mathbf{V}}$.
Then, attribute recognition for all instances previously found is achieved by aggregating the attribute fields over these instances.

First, we select cells $(x,y)$ whose confidence values $\mathbf{S}(x,y)$ are greater than a given threshold $\gamma$, in order to keep likely pedestrian cells only.
As in \cite{kendall2018multi}, the estimated centers $(x,y)+\vec{\mathbf{V}}(x,y)$ pointed at by the field $\vec{\mathbf{V}}$ from retained cells $(x,y)$ are then clustered into a set of instances with OPTICS algorithm~\cite{ankerst1999optics}, which has the property of not needing a prior number of clusters.
This yields $P$ clusters $\mathcal{C}(1), \ldots, \mathcal{C}(P)$, \ie, $P$ pedestrian detections, where each cluster $\mathcal{C}(p)$ contains all the cells $(x,y)$ associated with pedestrian $p$.

For each detected pedestrian $p$, associated with cluster $\mathcal{C}(p)$, we compute global predictions for all attributes based on the cells $(x,y)$ in the cluster.
First, the confidence score $s_p$ for the detection, used to rank detections for evaluation, is the average of scores $\mathbf{S}(x,y)$ over all cells:
\begin{equation}
    s_p = \operatorname{Sigmoid}\left( \frac{1}{\left| \mathcal{C}(p) \right|} \sum_{(x,y) \in \mathcal{C}(p)} \mathbf{S}(x,y) \right),
\end{equation}
where $\left| \mathcal{C}(p) \right|$ is the number of cells in the cluster.
Then, all other predictions are weighted averages with weights coming from the field $\mathbf{S}$.
A scalar (binary, categorical or continuous) attribute prediction $f_p$ is given by
\begin{equation}
    f_p = \operatorname{act}\left( \frac{\sum_{(x,y) \in \mathcal{C}(p)} \operatorname{Sigmoid}\left(\mathbf{S}(x,y)\right) \mathbf{F}(x,y)}
                                        {\sum_{(x,y) \in \mathcal{C}(p)} \operatorname{Sigmoid}\left(\mathbf{S}(x,y)\right)} \right),
\end{equation}
where $\operatorname{act}$ is an activation function used to get the right output range depending on the attribute type, as defined in \Cref{sec:field_learning}.
Note that for the categorical case, $\mathbf{F}(x,y)$ have values for all classes of the attributes.
A vectorial attribute prediction $\vec{f}_p$ is computed by
\begin{equation}
    \begin{split}
        \vec{f}_p &= (x^{f}_p,y^{f}_p) \\
    			  &= \frac{\sum_{(x,y) \in \mathcal{C}(p)} \operatorname{Sigmoid}\left(\mathbf{S}(x,y)\right) \left( (x,y) + \vec{\mathbf{F}}(x,y) \right)}
                          {\sum_{(x,y) \in \mathcal{C}(p)} \operatorname{Sigmoid}\left(\mathbf{S}(x,y)\right)}.
    \end{split}
\end{equation}

We use this architecture to jointly do pedestrian detection and attribute recognition for each of the detections.
However, scaling to numerous attributes in a multi-task framework is notoriously difficult~\cite{kokkinos2017ubernet}, and often leads to drops in performance with respect to the single-task models.
In the next section, we study this phenomenon more in depth for our model.

\section{Scaling Learning to Multiple Attributes}
\label{sec:mtl_norm}

We now analyse the behavior of the MTL network at the fork in the architecture, between the shared backbone and all the task-specific heads, when varying the number of tasks.
We first shed light on an issue about gradient scales, and then suggest a simple solution to fix it and stabilize training.

\subsection{MTL Gradient Scale Issue}

We consider the common MTL architecture for $T$ tasks, composed of a shared backbone branching into $T$ predictors.
For our model, this corresponds to one task for each field.
Each branch $t$ is associated to the corresponding task $t$ with loss function $\mathcal{L}_t$ and task weight $\lambda_t$.
For simplicity, we assume for now that all examples are annotated for all $T$ tasks, but we later describe a slight modification to apply when annotations are heterogeneous across examples, as is the case in this paper.
As usual, the loss function $\mathcal{L}$ for the whole model is the weighted sum of all branch losses:
\begin{equation}
    \mathcal{L} = \sum_{t=1}^{T} \lambda_t \mathcal{L}_t.
    \label{eq:loss_forward}
\end{equation}
During the backward pass, for a feature $z$ from the shared backbone, the gradient $d \mathcal{L}/d z$ can be bounded by the norms of all task-specific gradients at the same point:
\begin{equation}
    \left\lVert \frac{d \mathcal{L}}{d z} \right\rVert
        \leq \sum_{t=1}^{T} \left\lVert \lambda_t \frac{d \mathcal{L}_t}{d z} \right\rVert
        \leq T \max_{1 \leq t \leq T} \left\lVert \lambda_t \frac{d \mathcal{L}_t}{d z} \right\rVert.
    \label{eq:back_bound}
\end{equation}
Assuming the weights $\lambda_t$ are not normalized by $T$ (\eg, by forcing them to sum to 1), which is a common case~\cite{chen2018gradnorm, guo2018dynamic, kendall2018multi}, the bound on the norm directly grows with $T$.
Although this does not imply that the gradients will actually grow with the bound, we experimentally observe it, as shown in \Cref{fig:grad_norm}~\hyperlink{fig:grad_norm}{(a)}.
This exploding gradient phenomenon in the shared backbone could then lead to unstable training with too many tasks.

A simple solution to this issue could be to average the losses rather than summing them, \ie, normalizing the weights $\lambda_t$ by $T$ in \Cref{eq:loss_forward}, or, equivalently, decreasing the learning rate by a factor $T$.
The bound on the gradient then becomes
\begin{equation}
    \left\lVert \frac{d \mathcal{L}}{d z} \right\rVert
        \leq \frac{1}{T} \sum_{t=1}^{T} \left\lVert \lambda_t \frac{d \mathcal{L}_t}{d z} \right\rVert
        \leq \max_{1 \leq t \leq T} \left\lVert \lambda_t \frac{d \mathcal{L}_t}{d z} \right\rVert,
    \label{eq:grad_stable_bound}
\end{equation}
which is independent of $T$ and prevents the backbone gradients to explode (assuming all loss functions are stable), as observed in \Cref{fig:grad_norm}~\hyperlink{fig:grad_norm}{(b)}.
However, the effective loss weights $\lambda_{t}/T$ now decrease with $T$.
Since for a feature (including output) $z_t$ in a task-specific branch $t$ (\ie, after the fork), the loss $\mathcal{L}_t$ is the only one influenced by $z_t$, the gradient with respect to $z_t$ becomes
\begin{equation}
    \frac{d \mathcal{L}}{d z_t} = \frac{\lambda_t}{T} \frac{d \mathcal{L}_t}{d z_t}.
\end{equation}
The norm of this gradient now decreases with $T$ too, so that the task-specific branches learn slower with more tasks.

We interpret this as a difference in learning dynamics between the shared and task-specific sub-networks inherent to MTL, which should be more pronounced the more tasks there are.
This was already hinted at in UberNet~\cite{kokkinos2017ubernet}, which uses a different counter for each part (any task-specific predictor or the shared backbone) and updates them separately depending on the number of examples back-propagated through each of them.
Lu et al.~\cite{lu202012} also pointed out the importance to separate learning rates for the shared backbone and task-specific branches.

\subsection{Fork-Normalizing MTL Gradient Back-Propagation}

In order to get the best of both worlds, \ie, gradient bounds independent of the number of tasks $T$ both in the shared backbone and the task-specific heads, we propose to modify how gradients join at the fork during the backward pass.
We introduce a set of $T$ parameters $\kappa = \left( \kappa_1, \ldots, \kappa_T \right)$ to weigh the gradients coming from different branches in the merging, which we call fork-normalization.
Note that the forward pass (\Cref{eq:loss_forward}) is left unchanged, only the learning dynamic is adapted during the backward one, which now becomes:
\begin{equation}
    \begin{dcases}
        \frac{d \mathcal{L}}{d z}
            = \sum_{t=1}^{T} \kappa_t \lambda_t \frac{d \mathcal{L}_t}{d z}, \; \text{for $z$ in backbone,}\\
        \frac{d \mathcal{L}}{d z_t}
            = \lambda_t \frac{d \mathcal{L}_t}{d z_t}, \; \text{for $z_t$ in branch $t$}.
    \end{dcases}
\end{equation}

There are now two sets of parameters $\lambda_t$ and $\kappa_t$ associated with tasks $t$, but they have different purposes.
The first are used to balance the relative importances of tasks, to bias learning toward any of them.
The weights can be chosen by cross-validation, or tuned dynamically based on the uncertainty inherent to the tasks~\cite{kendall2018multi} or on the relative rate at which they learn~\cite{chen2018gradnorm, guo2018dynamic} for example.
On the other hand, the $\kappa$ parameters are used to balance learning in the shared backbone relative to the task-specific branches, and we propose strategies to choose them in this paper.

Several ways are possible for choosing $\kappa$.
The standard gradient accumulation formulation is obtained by setting $\kappa_t = 1$ for all $t$.
We first propose to choose $\kappa$ so that the coefficients sum to 1, \ie, $\sum_{t=1}^{T} \kappa_t = 1$.
As shown in \Cref{fig:grad_norm}~\hyperlink{fig:grad_norm}{(c)}, this yields a stable bound on the backbone gradient norm as in \Cref{eq:grad_stable_bound}, while still keeping the learning of the task-specific branches unchanged and independent of $T$.
For each example, we sample a $T$-tuple from a symmetric Dirichlet distribution $\kappa \sim \text{Dir}(\alpha)$ where all the $T$ dimensions have the same concentration parameter $\alpha$.
We study three particular cases of fork-normalization here:
\begin{enumerate}[label=(\roman*)]
    \item random, where $\kappa$ is sampled from the uniform distribution over the $T$-tuples ($\alpha = 1$);
    \item sample, where a single task $t$ is selected for back-propagation in the backbone with $\kappa_t = 1$ ($\alpha \rightarrow 0$);
    \item average, where all $\kappa_t = 1/T$ have the same constant value ($\alpha \rightarrow + \infty$).
\end{enumerate}

Although this approach yields a stable upper bound on the gradient norm (independent of $T$), we experimentally observe in \Cref{fig:grad_norm}~\hyperlink{fig:grad_norm}{(c)} that the actual norm decreases in the backbone when increasing the number of tasks $T$, meaning that the bound on gradient norm from \Cref{eq:back_bound} is not tight.
To address this, we model the relation between the gradient norm and the number of tasks as linear in log-log space, \ie, a power law.
This yields new gradient weighting parameters of the form $\kappa_t = 1/T^\beta$, where $\beta$ is a hyper-parameter of the power law.
From the measurement data in the graphs from \Cref{fig:grad_norm}, we obtain $\beta=0.5$ under this modeling.
Results for this gradient fork-power normalization with $\kappa_t = 1/\sqrt{T}$ are displayed in \Cref{fig:grad_norm}~\hyperlink{fig:grad_norm}{(d)}, where we observe that gradient norms in both the shared backbone and task-specific heads are now much more stable with respect to the number of tasks $T$.

By keeping gradients at the same scale regardless of the number of tasks, the training should be more stable, with more balance between shared backbone and task-specific branches.
Hyper-parameters such as learning rate should also generalize better to the addition of tasks, making cross-validation easier to carry out.
In addition, this modification does not incur any overhead.
In the case of a single task ($T=1$), all methods are completely equivalent to the standard gradient accumulation.

Until here, we assumed for simplicity that targets for all $T$ tasks were always available.
When this is not the case for a given example, \ie, it is annotated only for a subset of $\widetilde{T} < T$ tasks, we apply the same strategy with the effective number of back-propagated tasks $\widetilde{T}$ in the normalizing factor $\kappa$.
For example, the fork-power normalization weights become $\kappa_t = 1/\sqrt{\widetilde{T}}$.
Note that $\widetilde{T}$ can now be different across examples.

Gradient fork-average and fork-power normalizations should be equivalent to having separate learning rates for the shared backbone and the task-specific heads, and adapting them based on the number of tasks.
Adam~\cite{kingma2015adam} is an optimizer that maintains a different learning rate per parameter, and could therefore be able to achieve similar results automatically.
However, we experimentally find that the issue is still present with Adam.

\section{Experiments}

\begin{table*}[t]
  \caption{\textbf{Comparison between multi- and single-task networks} on JAAD val set with $A=32$ attributes. APs (\%) are shown for some attributes, mAP (\%) is the average of APs for all $A=32$ attributes. TtC stands for Time-to-Crossing.}
  \label{tab:mtl_stl}
  \centering
  \addtolength{\tabcolsep}{-.6pt}
  \addtolength{\cmidrulekern}{-.6pt}
  \begin{tabular}{lcccccccccccc}
    \toprule
    & & & Detection & \multicolumn{2}{c}{Intention} & \multicolumn{3}{c}{Behavior} & \multicolumn{3}{c}{Appearance} & All \\
    \cmidrule(lr){4-4} \cmidrule(lr){5-6} \cmidrule(lr){7-9} \cmidrule(lr){10-12} \cmidrule(lr){13-13}
    MTL Strategy
    & Networks
    & Memory
    & Pedestrian
    & Crossing
    & TtC
    & Looking
    & Walking
    & Front Pose
    & Age
    & Gender
    & Phone
    & mAP \\
    \midrule
    32 Networks & 32 ResNet-18's & 38.4GB & 65.1 & 58.9 & 22.3 & \textbf{35.7} & 27.2 & 46.2 & 13.4 & 30.2 & 31.0 & 36.6 \\
    \rowcolor{gray!15}
    MTL Baseline & 1 ResNet-50 & \textbf{1.6GB} & 66.3 & 56.5 & 21.9 & 34.4 & 28.4 & 45.3 & 21.1 & 32.1 & 32.2 & 36.6 \\
    PCGrad~\cite{yu2020gradient} & 1 ResNet-50 & \textbf{1.6GB} & 66.4 & 59.2 & \textbf{24.1} & 34.2 & 26.8 & 47.6 & 22.1 & \textbf{35.7} & 31.7 & 37.3 \\
    \rowcolor{gray!15}
    Fork-Norm. MTL & 1 ResNet-50 & \textbf{1.6GB} & \textbf{67.3} & \textbf{59.9} & 24.0 & 34.9 & \textbf{28.9} & \textbf{48.3} & \textbf{22.3} & 34.6 & \textbf{33.3} & \textbf{38.8} \\
    \bottomrule
  \end{tabular}
  \addtolength{\cmidrulekern}{.6pt}
  \addtolength{\tabcolsep}{.6pt}
\end{table*}

\begin{figure*}
    \centering
    \includegraphics[width=0.5\linewidth]{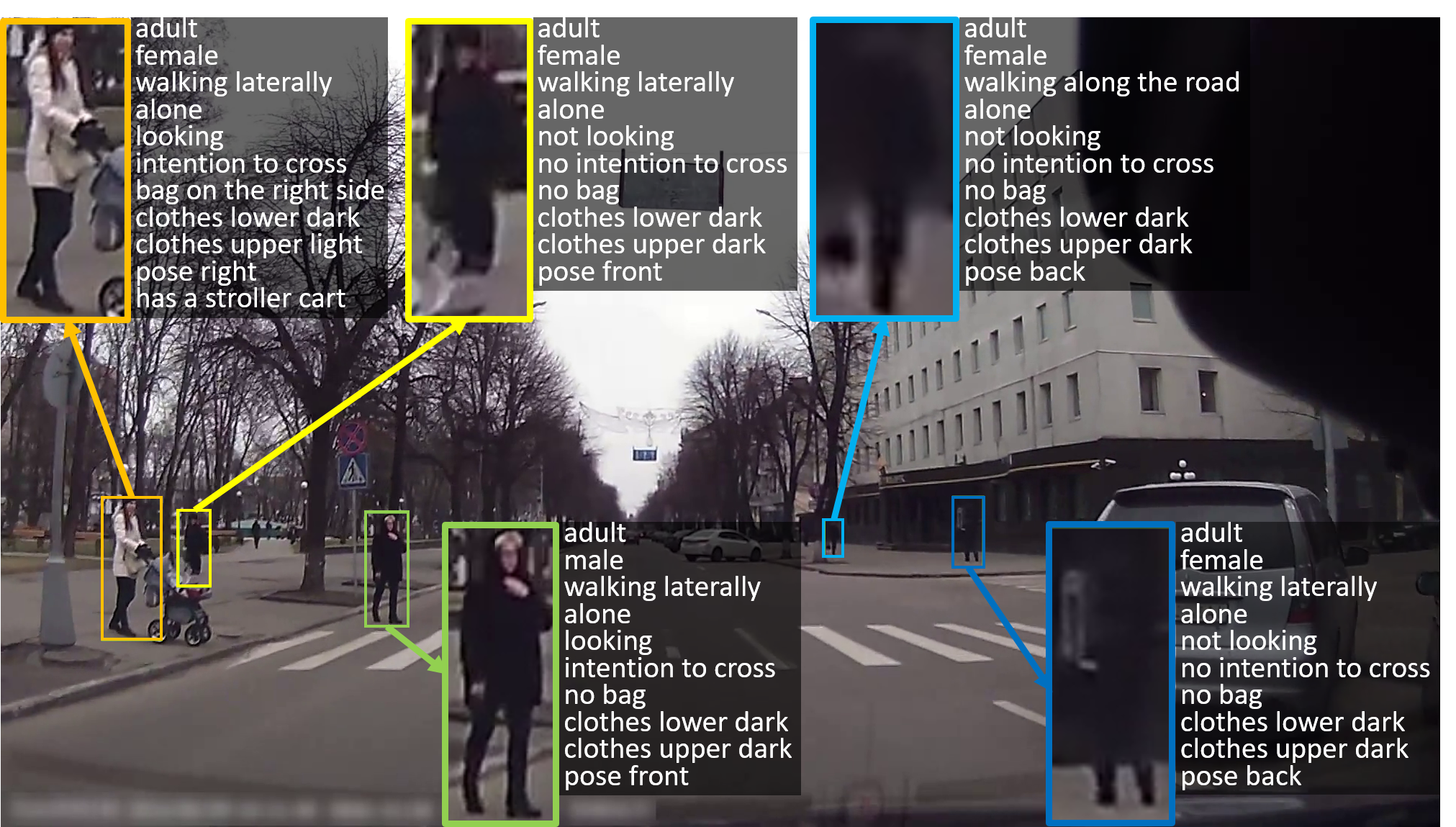}\includegraphics[width=0.5\linewidth]{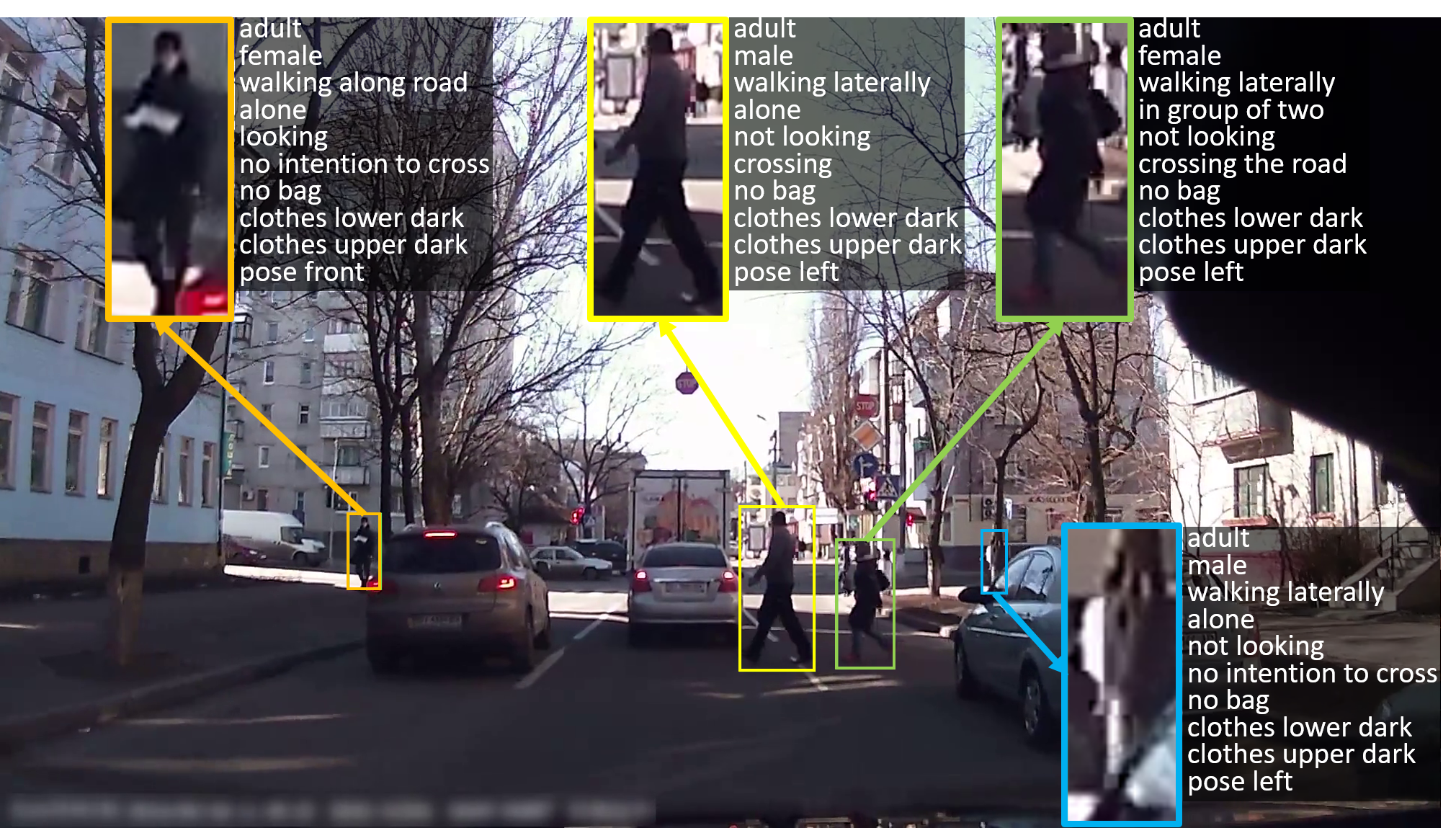}
    \caption{\textbf{Qualitative examples.} Results of the proposed pedestrian detection with multiple attributes on two scenes from JAAD val set. All pedestrians of sufficient apparent sizes are correctly detected as indicated by colored bounding boxes (5 in the left scene, 4 in the right scene). For each detected pedestrian, an inset zooms in the image region and indicates the detected attributes among: `age', `gender', `walking' and `motion direction', `group size', `looking', `road crossing intention' or `instant road crossing', `bag' (all types), 'upper clothing', `lower clothing', `back/front/left/right pose'. Difficult instances may not have ground truth annotation for some of the attributes, but the model still outputs plausible predictions.}
    \label{fig:ex1}
\end{figure*}

\begin{table*}[htb]
  \caption{\textbf{Evaluation of detection, box and image classification} on JAAD test set in accuracy (\%) and average precision (\%). Attributes learned by models are indicated in column `Attributes', a $\checkmark$ in column `Box' indicates that the method uses ground-truth bounding boxes as input, and a $\checkmark$ in column `Video' indicates that the method is trained and evaluated on video sequences. Metrics with a star are evaluated on balanced sets.}
  \label{tab:jaad_sota}
  \centering
  \addtolength{\tabcolsep}{-2.5pt}
  \addtolength{\cmidrulekern}{-2.5pt}
  \begin{tabular}{cllccccccccc}
    \toprule
    & \multicolumn{4}{c}{Model} & Detection & \multicolumn{4}{c}{Crossing} & Looking & Walking \\
    \cmidrule(lr){2-5} \cmidrule(lr){6-6} \cmidrule(lr){7-10} \cmidrule(lr){11-11} \cmidrule(lr){12-12}
    & Name
    & Attributes
    & Box
    & Video
    & Ped. AP
    & Box Acc.
    & Box AP
    & Img. Acc.
    & Img. AP
    & Box AP*
    & Box AP* \\
    \midrule
    (a) & RetinaNet~\cite{lin2017focal} (from \cite{pop2019multi}) & Det. & & & 56.1 & -- & -- & -- & -- & -- & -- \\
    \rowcolor{gray!15}
    (b) & Action+Context~\cite{rasouli2017are} & Cross.+Look.+Walk. & \checkmark & \checkmark & -- & -- & 62.7 & -- & -- & 80.2 & 83.5 \\
    (c) & & & & & -- & 81.0 & -- & -- & -- & -- & -- \\
    (d) & \multirow{-2}{*}{SKLT~\cite{fang2018is}} & \multirow{-2}{*}{Det.+Cross.} & & \checkmark & -- & 88.0 & -- & -- & -- & -- & -- \\
    \rowcolor{gray!15}
    (e) & & Det.+Cross. & & \checkmark & -- & -- & 73.8 & -- & -- & -- & -- \\
    \rowcolor{gray!15}
    (f) & \multirow{-2}{*}{ST-DenseNet~\cite{saleh2019real}} & Cross. & \checkmark & \checkmark & -- & -- & 84.8 & -- & -- & -- & -- \\
    (g) & Res-EnDec~\cite{gujjar2019classifying} & Cross. & & \checkmark & -- & -- & -- & 67.4 & 81.1 & -- & -- \\
    \rowcolor{gray!15}
    (h) & LookingAhead~\cite{chaabane2019looking} & Cross. & & \checkmark & -- & -- & -- & -- & 86.7 & -- & -- \\
    (i) & MTL Baseline & Det.+All Att. ($A=32$) & & & 70.0 & 86.5 & 82.3 & 78.3 & 92.9 & 75.6 & 88.1 \\
    \midrule
    \rowcolor{gray!15}
    (j) & Gradient Fork-Power Norm. & Det.+All Att. ($A=32$) & & & 70.4 & 87.4 & 83.8 & 80.6 & 94.3 & 76.8 & 88.8 \\
    (k) & Gradient Fork-Power Norm. & Det.+Cross.+Look.+Walk. & & & 71.0 & 87.2 & 82.7 & 81.8 & 94.6 & 82.8 & 87.3 \\
    \rowcolor{gray!15}
    (l) & Gradient Fork-Power Norm. & Det.+Cross. & & & 70.8 & 87.0 & 83.1 & 82.9 & 94.9 & -- & -- \\
    (m) & Gradient Fork-Power Norm. & Det.+Look. & & & 71.6 & -- & -- & -- & -- & 82.0 & -- \\
    \rowcolor{gray!15}
    (n) & Gradient Fork-Power Norm. & Det.+Walk. & & & 71.1 & -- & -- & -- & -- & -- & 85.9 \\
    (o) & Gradient Fork-Power Norm. & Det. & & & 71.1 & -- & -- & -- & -- & -- & -- \\
    \bottomrule
  \end{tabular}
  \addtolength{\cmidrulekern}{2.5pt}
  \addtolength{\tabcolsep}{2.5pt}
\end{table*}

\subsection{Experimental Setup}
\label{sec:exp_setup}

\paragraph{Dataset and Attributes}
We use JAAD dataset~\cite{rasouli2017are}, which is centered around pedestrian analysis from vehicles.
To the best of our knowledge, it is the only dataset where pedestrians are annotated both for detection and with a large number of diverse attributes.
We consider the default split, composed of 40,530 images (177 videos) for training, 7,170 images (29 videos) for validation, and 27,912 images (117 videos) for testing.
We extract 32 pedestrian attributes in total, which we divide in four sets of $A$ attributes:
\begin{enumerate}[label=(\alph*)]
    \item Detection only, \ie, bounding box attribute ($A=1$), but note that it corresponds to four task losses associated to fields $\mathbf{S}$, $\vec{\mathbf{V}}$, $\mathbf{F}_{h}$ and $\mathbf{F}_{w}$ as explained in \Cref{sec:model};
    \item Future intention attributes ($A=2$):
        \begin{itemize}
            \item binary: `road crossing intention';
            \item continuous: `time-to-crossing';
        \end{itemize}
    \item Current behavior attributes ($A=10$):
        \begin{itemize}
            \item binary: `instant road crossing', `looking', `walking', `motion direction', `back pose', `front pose', `left pose' and `right pose';
            \item categorical: `group size' and `reaction';
        \end{itemize}
    \item Appearance attributes ($A=19$):
        \begin{itemize}
            \item binary: `gender', `backpack', `bag at elbow', `bag at hand', `bag on left side', `bag on right side', `bag on shoulder', `cap', `clothes below knee', `dark lower clothes', `dark upper clothes', `light lower clothes', `light upper clothes', `hood', `object', `phone', `stroller cart' and `sunglasses';
            \item categorical: `age'.
        \end{itemize}
\end{enumerate}
We left out attributes about `hand gesture', `nod', `baby', `bicycle/motorcycle' and `umbrella' because there are too few examples to learn from for these attributes.
Note that all attributes are not available on all pedestrians, making the annotations heterogeneous across examples.
We also do not train on pedestrians too heavily occluded (more than 75\% occlusion).
More details can be found in the reference paper~\cite{rasouli2017are}.

In order to analyze the behavior of our model when varying the number of tasks, we train and evaluate on four increasing sets of attributes:
\begin{itemize}
    \item (a) with $A=1$;
    \item (a+b) with $A=3$;
    \item (a+b+c) with $A=13$;
    \item (a+b+c+d) with $A=32$.
\end{itemize}

\paragraph{Implementation Details}
Our model is based on a ResNet-50 backbone~\cite{he2016deep}, with single $1 \times 1$ sub-pixel convolution layers~\cite{shi2016real} as task-specific predictors.
We use pre-trained weights from PifPaf~\cite{kreiss2019pifpaf} since it uses a similar framework and is trained on humans specifically.
The loss functions are (binary) focal cross-entropy~\cite{lin2017focal} for (binary) classification tasks, and $L_1$ for regression ones (continuous scalar and vectorial attributes).

The networks are trained with SGD optimizer with a batch size of 4, learning rate of $5\cdot10^{-4}$, weight decay of $5\cdot10^{-4}$, momentum of $0.95$, and exponential model averaging with decay constant of $10^{-3}$.
The number of epochs is selected based on convergence for each variant and group of attributes separately.
Note that the results seem stable when changing the batch size, and that the standard gradient accumulation variant with all tasks ($A=32$) uses a learning rate halved to avoid unstable training due to exploding gradients.
Task weights are learned using uncertainty~\cite{kendall2018multi} with bigger values for detection tasks to bias toward them as they are the most important ones (factors of 2 for $A=3$ and $A=4$, 4 for $A=13$ and 7 for $A=32$).
The vote step uses threshold $\gamma=0.2$, and OPTICS~\cite{ankerst1999optics} with a minimum cluster size of 10, a maximum radius of 5, and cluster threshold of 0.5.

During training, images are augmented with random horizontal flipping, scaling to width 961px, cropping out of top third (making them 369px high), zoom in or out with a factor in [0.95, 1.05], random color jittering, random jpeg compression and random grayscale conversion.
This yields final feature maps of size $121 \times 47$ neurons with our network.

\paragraph{Detection and Classification Metrics}
We evaluate pedestrian detection with average precision (AP) at an Intersection-over-Union (IoU) threshold of 0.5.
Each binary or categorical attribute is evaluated by computing similar APs considering all values of the attribute as different classes of pedestrians and taking the mean over them.
For attribute \textit{time-to-crossing}, we consider APs for 10 classes corresponding to an absolute error lower than 10 thresholds, varying from 0.5s to 5s by step of 0.5s, and average them again.
A global mean AP (mAP) summarizing the performances across all attributes (depending on the set of tasks learned) is given by averaging detection AP and all attribute APs.

In order to enable comparison with box classification approaches (\ie, using ground-truth bounding boxes as input), we adapt our evaluation pipeline to fit this \textit{classification} setup.
Each ground-truth pedestrian is matched with the closest detection whose center is inside the bounding box, and the attribute predictions are taken from the matched detection.
When no match is found, predictions are the classes with the most examples from the training set.
Image-wise predictions for crossing forecasting are obtained from the detections with highest confidences on this attribute.
From our understanding, competing methods either use ground truth boxes as input or evaluate only on the set of correctly detected instances, but the results are then heavily dependent on the detection performance.
For the attributes \textit{looking} and \textit{walking}, the evaluation protocol of Rasouli et al.~\cite{rasouli2017are} is rather different, so we adopt a similar one.
For each of these two attributes, results are computed on a balanced test set, where all ground truths of the class with fewer examples are kept, and the same number of ground truths are randomly sampled from the other class.
Results are computed on predictions from the selected ground truths only.
The sampling is done 10 times and APs are averaged.
We denote these metrics with a star.

\begin{table*}[t]
  \caption{\textbf{Effect of gradient merging} on JAAD val set. APs (\%) are shown for some attributes, mAP (\%) is the average of APs for all attributes evaluated (different for each group). TtC stands for Time-to-Crossing.}
  \label{tab:ablation_gradmerge}
  \centering
  \begin{tabular}{lcccccccccc}
    \toprule
    & Detection & \multicolumn{2}{c}{Intention} & \multicolumn{3}{c}{Behavior} & \multicolumn{3}{c}{Appearance} & All \\
    \cmidrule(lr){2-2} \cmidrule(lr){3-4} \cmidrule(lr){5-7} \cmidrule(lr){8-10} \cmidrule(lr){11-11}
    Gradient Merging
    & Pedestrian
    & Crossing
    & TtC
    & Looking
    & Walking
    & Front Pose
    & Age
    & Gender
    & Phone
    & mAP \\
    \midrule
    \multicolumn{11}{c}{Detection Only ($A=1$)} \\
    \midrule
    Accumulation & 68.1 & -- & -- & -- & -- & -- & -- & -- & -- & 68.1 \\
    \rowcolor{gray!15}
    Mean Loss & 67.5 & -- & -- & -- & -- & -- & -- & -- & -- & 67.5 \\
    \midrule
    Fork-Sample Norm. & 66.8 & -- & -- & -- & -- & -- & -- & -- & -- & 66.8 \\
    \rowcolor{gray!15}
    Fork-Random Norm. & 67.5 & -- & -- & -- & -- & -- & -- & -- & -- & 67.5 \\
    Fork-Average Norm. & 67.3 & -- & -- & -- & -- & -- & -- & -- & -- & 67.3 \\
    \rowcolor{gray!15}
    Fork-Power Norm. & \textbf{68.2} & -- & -- & -- & -- & -- & -- & -- & -- & \textbf{68.2} \\
    \midrule
    \multicolumn{11}{c}{Detection + Intention Attributes ($A=3$)} \\
    \midrule
    Accumulation & 67.1 & 58.6 & \textbf{23.7} & -- & -- & -- & -- & -- & -- & 49.8 \\
    \rowcolor{gray!15}
    Mean Loss & 66.1 & 57.4 & 23.4 & -- & -- & -- & -- & -- & -- & 49.0 \\
    \midrule
    Fork-Sample Norm. & 66.6 & 57.8 & 23.3 & -- & -- & -- & -- & -- & -- & 49.2 \\
    \rowcolor{gray!15}
    Fork-Random Norm. & 66.9 & 59.8 & 23.4 & -- & -- & -- & -- & -- & -- & 50.0 \\
    Fork-Average Norm. & 66.8 & 60.5 & \textbf{23.7} & -- & -- & -- & -- & -- & -- & \textbf{50.3} \\
    \rowcolor{gray!15}
    Fork-Power Norm. & \textbf{67.2} & \textbf{60.6} & 22.9 & -- & -- & -- & -- & -- & -- & 50.2 \\
    \midrule
    \multicolumn{11}{c}{Detection + Intention and Behavior Attributes ($A=13$)} \\
    \midrule
    Accumulation & 65.0 & 58.0 & 22.0 & 33.2 & \textbf{29.9} & 45.3 & -- & -- & -- & 39.2 \\
    \rowcolor{gray!15}
    Mean Loss & 65.5 & 56.4 & 22.3 & 31.3 & 26.8 & 42.8 & -- & -- & -- & 38.4 \\
    \midrule
    Fork-Sample Norm. & 60.1 & 53.4 & 20.4 & 30.6 & 25.2 & 40.0 & -- & -- & -- & 35.9 \\
    \rowcolor{gray!15}
    Fork-Random Norm. & 65.3 & 59.0 & 23.5 & 31.7 & 29.0 & 46.5 & -- & -- & -- & 40.1 \\
    Fork-Average Norm. & 65.7 & 59.6 & \textbf{23.7} & 32.4 & 29.4 & 48.8 & -- & -- & -- & 40.7 \\
    \rowcolor{gray!15}
    Fork-Power Norm. & \textbf{66.9} & \textbf{60.2} & 23.6 & \textbf{33.7} & \textbf{29.9} & \textbf{49.3} & -- & -- & -- & \textbf{41.4} \\
    \midrule
    \multicolumn{11}{c}{Detection + Intention, Behavior and Appearance Attributes ($A=32$)} \\
    \midrule
    Accumulation & 66.3 & 56.5 & 21.9 & 34.4 & 28.4 & 45.3 & 21.1 & 32.1 & 32.2 & 36.6 \\
    \rowcolor{gray!15}
    Mean Loss & 66.3 & 59.6 & 22.6 & 32.7 & 27.6 & \textbf{48.3} & 19.8 & 30.4 & 31.9 & 35.9 \\
    \midrule
    Fork-Sample Norm. & 54.3 & 49.3 & 16.9 & 28.0 & 24.2 & 38.7 & 13.6 & 24.4 & 26.9 & 30.5 \\
    \rowcolor{gray!15}
    Fork-Random Norm. & 65.4 & 59.4 & 23.5 & 32.2 & 29.4 & 47.4 & 21.9 & 31.3 & 31.5 & 36.7 \\
    Fork-Average Norm. & 65.5 & 58.8 & 22.9 & 32.6 & 28.1 & 46.3 & 17.0 & 27.6 & 31.7 & 35.5 \\
    \rowcolor{gray!15}
    Fork-Power Norm. & \textbf{67.3} & \textbf{59.9} & \textbf{24.0} & \textbf{34.9} & \textbf{28.9} & \textbf{48.3} & \textbf{22.3} & \textbf{34.6} & \textbf{33.3} & \textbf{38.8} \\
    \bottomrule
  \end{tabular}
\end{table*}

\subsection{Motivation for Multi-Task Learning}

In an autonomous vehicle, on-board hardware sets strict limits on the memory, the number of operations, the power consumption and the inference time for the models.
A fair comparison should therefore take these constraints into account in addition to the performances. 
Although results may change when the number of attributes learned increases, this would have to be balanced with run-time resource utilization in practice.
In order to motivate a MTL approach, we compare in \Cref{tab:mtl_stl} our MTL network to a collection of single-task networks, with one model learned for each attribute.
In this section, learning is done on the training set and evaluation on the validation set.

Our MTL model takes 1.6GB of memory to predict all 32 attributes.
This means that, on average, multiple single-task networks should take no more than 50MB each, in order to equate the memory footprint of both approaches, which would yield very small networks and probably poor results.
For quantitative comparison, we trained the collection of models with ResNet-18 as backbone networks, \ie, the smallest network available that still allows the same experimental setup, in particular, which has PifPaf pre-trained weights.
Although this approach uses smaller networks, it is important to note that it still has a memory footprint of 38.4GB, \ie, 24 times bigger than our MTL model.
Aside from memory, it does on par with the MTL baseline, \ie, our model without fork-normalization (standard accumulation).
However, our fork-normalization approach outperforms the collection of networks, with both less memory and better attribute APs.
The results thus validate the use of MTL in embedded applications with run-time requirements.

We also compare with PCGrad~\cite{yu2020gradient}, another contemporary approach from the literature to stabilize training and prevent negative transfer between tasks.
PCGrad computes gradients for all tasks separately, and modifies them when pairs of gradients have negative scalar products, so that all gradients point in more similar directions.
It is noticeable that PCGrad requires a separate backward pass for each task, which does not scale well to numerous tasks in terms of training time, while fork-normalization only happens when joining gradients at the fork and does not bring overhead.
We implement PCGrad on top of our MTL baseline model.
Although it also improves on the baseline and the collection of networks, it is outperformed by our fork-normalization approach.

Two examples of predictions from our model are displayed in \Cref{fig:ex1}.
Overall, we observe that detections are accurately localized, with no or few false positives.
Although qualitative evaluation of attributes is harder due to low resolution of most pedestrians, the predictions look generally convincing.

\begin{table*}[t]
  \caption{\textbf{Impact of number of tasks} on JAAD val set. APs (\%) are given for some attributes when learned only with detection ($A=2$) or along with all attributes ($A=32$), and AP gaps when scaling to all attributes are indicated to show the evolutions of performances.}
  \label{tab:ablation_number_tasks}
  \centering
  \begin{tabular}{lccc}
    \toprule
    Gradient Merging
    & \begin{tabular}{@{}c@{}}Detection and\\Single Attribute\\($A=2$)\end{tabular}
    & \begin{tabular}{@{}c@{}}Detection and\\All Attributes\\($A=32$)\end{tabular}
    & AP Gap \\
    \midrule
    \multicolumn{4}{c}{Crossing} \\
    \midrule
    Accumulation & 58.1 & 56.5 & -1.6 (-2.8\%) \\
    \rowcolor{gray!15}
    Fork-Power Norm. & 58.1 & 59.9 & \textbf{+1.8} (\textbf{+3.1\%}) \\
    \midrule
    \multicolumn{4}{c}{Looking} \\
    \midrule
    Accumulation & 34.0 & 34.4 & +0.4 (+1.2\%) \\
    \rowcolor{gray!15}
    Fork-Power Norm. & 34.2 & 34.9 & \textbf{+0.7} (\textbf{+2.0\%}) \\
    \midrule
    \multicolumn{4}{c}{Front Pose} \\
    \midrule
    Accumulation & 47.9 & 45.3 & -2.6 (-5.4\%) \\
    \rowcolor{gray!15}
    Fork-Power Norm. & 45.5 & 48.3 & \textbf{+2.8} (\textbf{+6.2\%}) \\
    \midrule
    \multicolumn{4}{c}{Gender} \\
    \midrule
    Accumulation & 32.1 & 32.1 & 0.0 (0.0\%) \\
    \rowcolor{gray!15}
    Fork-Power Norm. & 32.7 & 34.6 & \textbf{+1.9} (\textbf{+5.8\%}) \\
    \bottomrule
  \end{tabular}
  \hfill
  \begin{tabular}{lccc}
    \toprule
    Gradient Merging
    & \begin{tabular}{@{}c@{}}Detection and\\Single Attribute\\($A=2$)\end{tabular}
    & \begin{tabular}{@{}c@{}}Detection and\\All Attributes\\($A=32$)\end{tabular}
    & AP Gap \\
    \midrule
    \multicolumn{4}{c}{Time-to-Crossing} \\
    \midrule
    Accumulation & 23.6 & 21.9 & -1.7 (-7.2\%) \\
    \rowcolor{gray!15}
    Fork-Power Norm. & 24.1 & 24.0 & \textbf{-0.1} (\textbf{-0.4\%}) \\
    \midrule
    \multicolumn{4}{c}{Walking} \\
    \midrule
    Accumulation & 26.2 & 28.4 & +2.2 (+8.4\%) \\
    \rowcolor{gray!15}
    Fork-Power Norm. & 26.0 & 28.9 & \textbf{+2.9} (\textbf{+11.2\%}) \\
    \midrule
    \multicolumn{4}{c}{Age} \\
    \midrule
    Accumulation & 20.6 & 21.1 & +0.5 (+2.4\%) \\
    \rowcolor{gray!15}
    Fork-Power Norm. & 18.8 & 22.3 & \textbf{+3.5} (\textbf{+18.6\%}) \\
    \midrule
    \multicolumn{4}{c}{Phone} \\
    \midrule
    Accumulation & 34.4 & 32.2 & -2.2 (-6.4\%) \\
    \rowcolor{gray!15}
    Fork-Power Norm. & 32.6 & 33.3 & \textbf{+0.7} (\textbf{+2.1\%}) \\
    \bottomrule
  \end{tabular}
\end{table*}

\subsection{Comparison with the State of the Art}
\label{sec:sota}

Comparisons with approaches from the literature are presented in \Cref{tab:jaad_sota} for detection and classification on attributes \textit{crossing} (\ie, whether the pedestrian will cross the road in front of the vehicle), \textit{looking} (\ie, eye contact with the vehicle), and \textit{walking} (\ie, standing or walking posture).
To the best of our knowledge, there is no other work jointly addressing both detection and attribute recognition that we could compare to.
For this reason, we adapt our evaluation to match as closely as possible what is done in other works, but note that inputs are sometimes different between models (\eg, images or videos), so results are not always exactly comparable.
In this section, learning is done on the union of the training and validation sets, and evaluation on the testing set.

Pedestrian detection results are given with Pedestrian AP in the column Detection of \Cref{tab:jaad_sota}.
To the best of our knowledge, only RetinaNet~\cite{lin2017focal} (a) has been used for pedestrian detection on JAAD~\cite{pop2019multi}, and it is outperformed by all our models by significant margins, with more than 25\% of improvement, validating our approach for detection.
The reason is that field-based methods are well suited for low resolution scenarios, such as for autonomous vehicles, thanks to the spatial aggregation post-processing naturally leveraging context~\cite{kreiss2019pifpaf}.
On the other hand, RetinaNet relies on prior box classification, which is limited in the case of pedestrians of small apparent sizes.

\Cref{tab:jaad_sota} also presents the performances for the three attributes evaluated by previous works in the remaining columns.
Our approach either outperforms or performs on par with state-of-the-art methods that use ground-truth detections and/or videos, while still learning multiple additional attributes.
On attribute \textit{crossing}, SKLT~\cite{fang2018is} (c-d) slightly outperforms our method (j) by 0.7\% when using video.
However, on single frames as we do, it falls behind, 7.3\% worse when compared to our model.
When ST-DenseNet~\cite{saleh2019real} (e-f) uses predicted boxes, we outperform it by 13.6\%, even without video information.
Even their best result with ground-truth boxes is only 1.2\% better than ours.
Fork-normalization also compares favorably to the state of the art (g-h) on image-wise metrics.
Regarding attributes \textit{looking} and \textit{walking}, we outperform Action+Context~\cite{rasouli2017are} (b) when using a similar evaluation protocol on balanced sets.
Overall, our gradient fork-power normalization (j) better generalizes to many tasks than standard accumulation (i).
It is noticeable that our approach is generic regarding the choice of attributes, and can predict any kind of attributes, while competing models are designed and optimized for the specific attributes they predict.

\subsection{Ablation Study}
\label{sec:ablation}

We carry out several ablation studies to analyze the impact of the gradient merging operation depending on the number of tasks learned.
In this section, learning is done on the training set and evaluation on the validation set.

\Cref{tab:ablation_gradmerge} summarizes results of all the gradient merging variants on the four sets of tasks.
Note that the mAP scores cannot be compared between different sets of attributes as the averages do not cover the same numbers of values.
Overall, the fork-sample version clearly yields the worse results among all methods, possibly due to the fact that only one task is considered per example for each backward pass, therefore discarding a lot of supervision and introducing some noise.
The mean loss approach also underperforms, although the gap is not as large.
Among the four remaining methods, the gradient fork-normalization variants generally improve on the standard gradient accumulation, and the fork-power normalization consistently obtains the best global mAP results, \ie, over all attributes, or very close.
Moreover, the gaps with the standard gradient accumulation get bigger the more tasks are learned, from 0.1\% to 6.0\%.
Note that the learning rate for gradient accumulation trained on all tasks is halved because of exploding gradient issues.

\Cref{tab:ablation_number_tasks} investigates the AP gap from learning a single attribute to all of them.
Our proposed approach consistently sees lesser drops or higher gains, and is less sensitive to the number of tasks thanks to more stable gradients.
These results confirm that normalizing gradients at the fork helps scaling to numerous tasks.

\section{Conclusions}

We introduced a Multi-Task Learning (MTL) approach for joint pedestrian detection and attribute recognition.
It relies on a bottom-up field formalism, particularly suited to the low resolution context of autonomous vehicles.
We experimented with detection of up to 32 pedestrian attributes simultaneously. Our final model detecting 32 attributes outperforms the state-of-the-art (RetinaNet~\cite{lin2017focal}) single-task pedestrian detection by more than 25\%.
By increasing the number of attributes learned by the network, we highlighted an issue linked to gradient scale in MTL with numerous tasks.
We solved it by normalizing back-propagation at the fork in the architecture, leading to a more stable training and better generalization to the addition of tasks.
Although we only show results for this model in the context of pedestrian analysis, we think this approach can have applications in more general MTL frameworks, as the reasoning about gradient norms should be generic.
Furthermore, all attributes share the same common backbone.
This is a standard architecture for Multi-Task Learning, but it has the drawback of being independent of the specific attributes learned.
However, attributes may be closely related or completely different from each other, which would require different levels of sharing between them.
As future work, we would like to study the impact of sharing and its relation to the natures of attributes, depending for example on their high-level categories (e.g., intentions, behaviors, appearances).

Over the past years, academics and industry have joined forces to make autonomous vehicles a reality and save human lives.
While many challenges remain, we believe that pedestrian safety must be one of the highest priorities. For many years, researchers focused on simply detecting them.
In this work, we argue that we need to go beyond simple detection and detect as many attributes as possible from a single images.
These attributes will help better anticipate pedestrian behaviors.
However, our model performs a simple inference on future behaviors, and no further modeling of the pedestrians’ behaviors or past trajectories is done explicitly.
These aspects could only be represented implicitly through the internal features learned by the network, so combining our model with other complementary approaches would be a way to address this limitation.
In this situation, borrowing ideas from Traffic Psychology could help improve the predictions, with, for example, a dedicated module inferring pedestrians’ intentions.
It is also noticeable that the behavior of the ego vehicle can affect pedestrians’ decisions, therefore interactions with road users should also be taken into account.
While our work will impact the safety of autonomous vehicles, it can also be used in any Advanced Driver Assistance Systems (ADAS).
Anticipating early enough whether a pedestrian will cross the road will potentially save many lives.
We hope that our work will foster more research in this area.

\section{Acknowledgements}

We would like to thank  Valeo for funding our work, Sven Kreiss for sharing the code for field formalism (PifPaf~\cite{kreiss2019pifpaf}), and reviewers for their helpful comments.

\IEEEtriggercmd{\addtolength{\textheight}{-10cm}}
\IEEEtriggeratref{40}

\bibliography{biblio}

\begin{IEEEbiography}[{\includegraphics[width=1in,height=1.25in,clip,keepaspectratio]{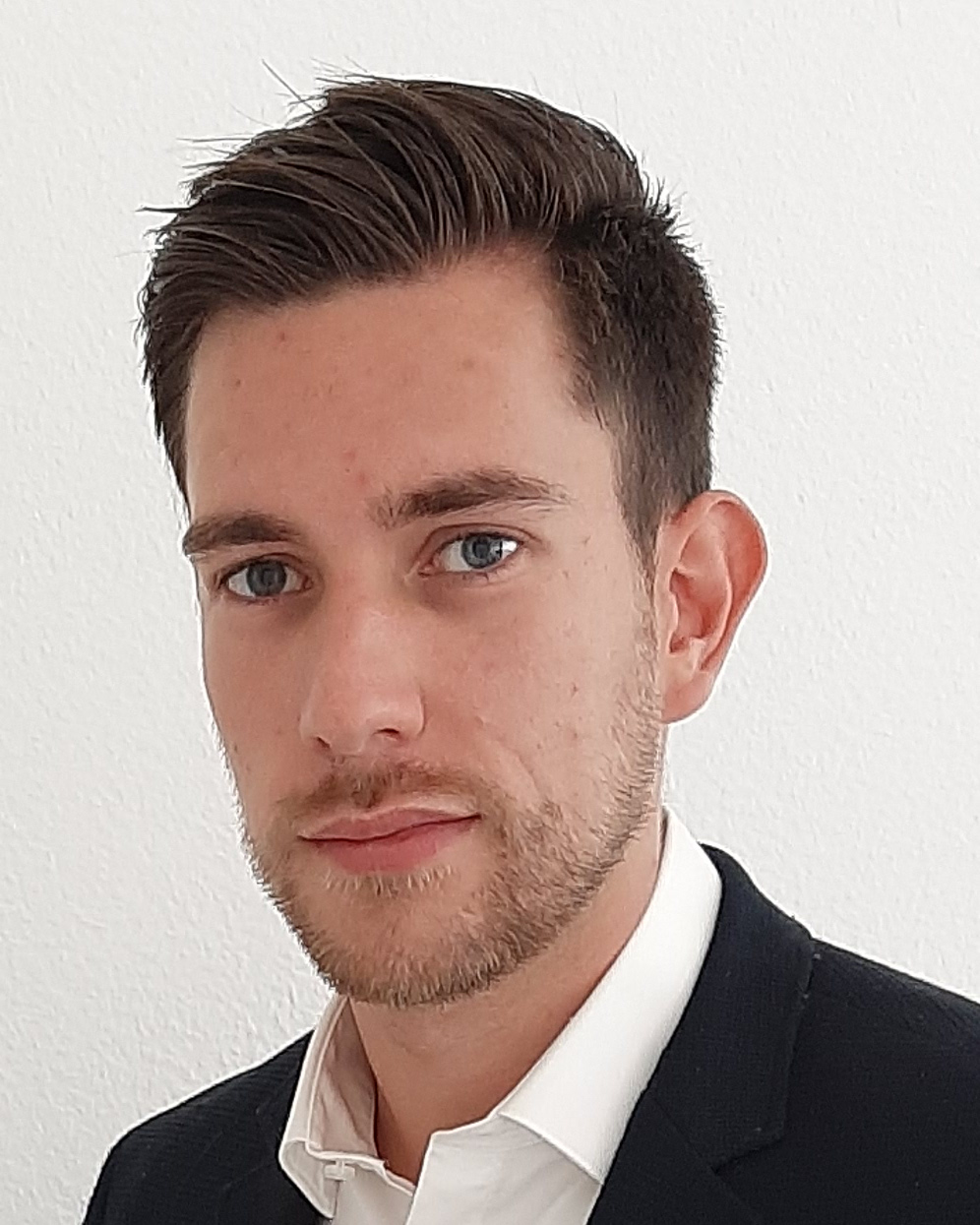}}]%
{Taylor Mordan}
received the engineering degree from ENSTA ParisTech, Paris, France, and the M.S. degree in computer science from UPMC, Paris, France, in 2015, then the Ph.D. degree in computer science from Sorbonne University, Paris, France, in 2018.
From 2015 to 2018, he was a Research Assistant with Thales LAS France. Since 2019, he has been a Post-Doctoral Researcher with VITA lab, EPFL, Lausanne, Switzerland. His research interests include computer vision, multi-task learning, and applications to perception in autonomous vehicles.
\end{IEEEbiography}

\begin{IEEEbiography}[{\includegraphics[width=1in,height=1.25in,clip,keepaspectratio]{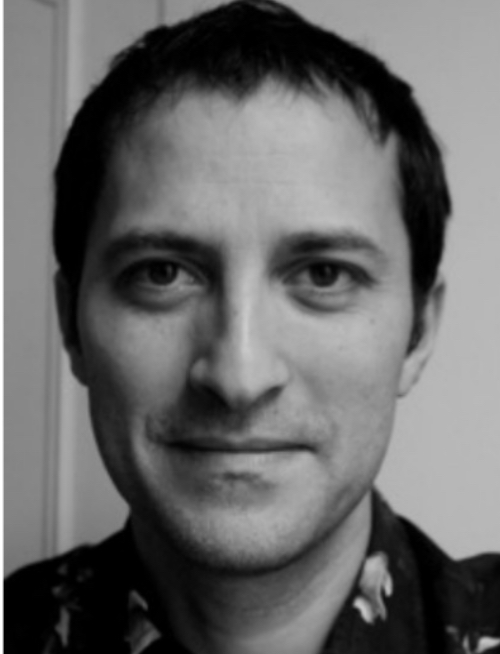}}]%
{Matthieu Cord}
is full professor at Sorbonne University. He is also part-time principal scientist at Valeo.ai. His research expertise includes computer vision, machine learning and artificial intelligence. He is the author of more 150 publications on image classification, segmentation, deep learning, and multimodal vision and language understanding. He is an honorary member of the Institut Universitaire de France and served from 2015 to 2018 as an AI expert at CNRS and ANR (National Research Agency).
\end{IEEEbiography}

\begin{IEEEbiography}[{\includegraphics[width=1in,height=1.25in,clip,keepaspectratio]{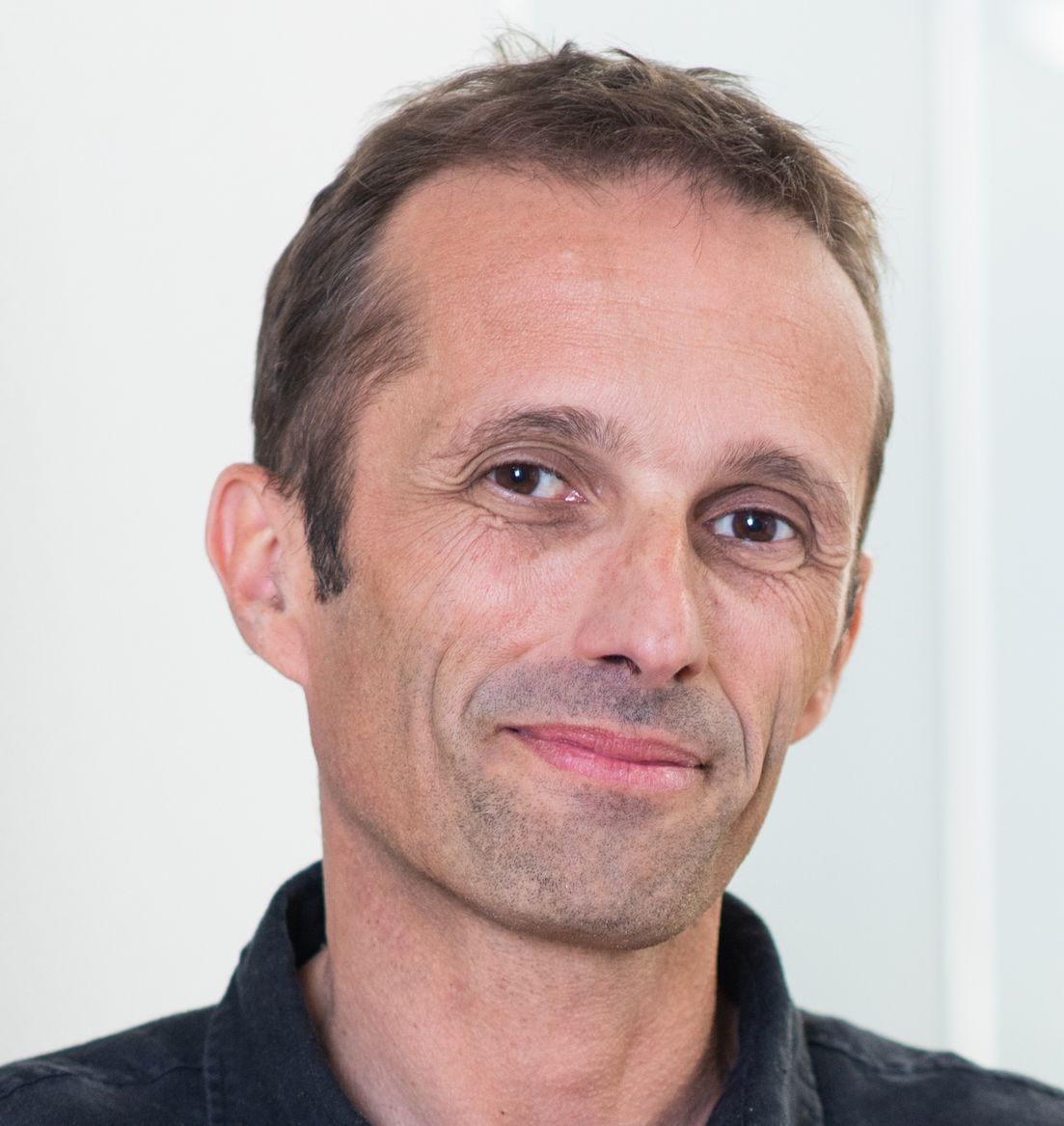}}]%
{Patrick Pérez}
is Scientific Director of Valeo.ai, a Valeo research lab on artificial intelligence for automotive applications. Before joining Valeo, Patrick Pérez has been Distinguished Scientist at Technicolor (2009-2918), researcher at Inria (1993-2000, 2004-2009) and at Microsoft Research Cambridge (2000-2004). His research revolves around machine learning for scene understanding, data mining and visual editing.
\end{IEEEbiography}

\begin{IEEEbiography}[{\includegraphics[width=1in,height=1.25in,clip,keepaspectratio]{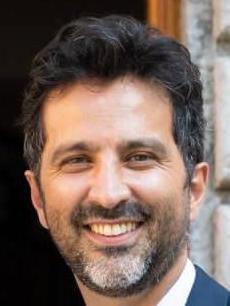}}]%
{Alexandre Alahi}
is currently an Assistant Professor at EPFL. He spent five years at Stanford University as a Post-doc and Research Scientist after obtaining his Ph.D. from EPFL. His research enables machines to perceive the world and make decisions in the context of transportation problems and smart environments. He has worked on the theoretical challenges and practical applications of socially-aware Artificial Intelligence, i.e., systems equipped with perception and social intelligence. 
\end{IEEEbiography}

\end{document}